\documentclass[lettersize,journal]{IEEEtran}
\usepackage{amsmath,amsfonts}
\usepackage[lined,boxed,commentsnumbered,ruled]{algorithm2e}
\usepackage{array}
\usepackage{cite}
\usepackage{subfigure}
\usepackage{graphicx}
\usepackage{ragged2e} 
\usepackage{booktabs,makecell, multirow, tabularx}
\usepackage{makecell}
\usepackage{color}
\usepackage{url}
\newtheorem{myDef}{Definition}
\newcommand\mymodel{HAGNN}
\begin{document}

\title{HAGNN: Hybrid Aggregation for Heterogeneous Graph Neural Networks}

\author{Guanghui~Zhu, ~\IEEEmembership{Member,~IEEE,}
	    Zhennan~Zhu,
        Hongyang Chen, ~\IEEEmembership{Member,~IEEE,}
        Chunfeng~Yuan,
        and~Yihua~Huang
\thanks{Guanghui~Zhu, Zhennan~Zhu, Chunfeng~Yuan, and Yihua~Huang are with State Key Laboratory for Novel Software Technology, Nanjing University, Nanjing, China. E-mail: zhuzhennan@smail.nju.edu.cn, \{zgh, cfyuan, yhuang\}@nju.edu.cn. Hongyang Chen is with the Research Center for Graph Computing, Zhejiang Lab, Hangzhou, China. E-mail: dr.h.chen@ieee.org. Guanghui Zhu is the corresponding author.}
\thanks{Manuscript received April 19, 2021; revised August 16, 2021.}}

\markboth{Journal of \LaTeX\ Class Files,~Vol.~14, No.~8, August~2021}%
{Shell \MakeLowercase{\textit{et al.}}: A Sample Article Using IEEEtran.cls for IEEE Journals}


\maketitle

\begin{abstract}
%
Heterogeneous graph neural networks (GNNs) have been successful in handling heterogeneous graphs.
In existing heterogeneous GNNs, meta-path plays an essential role.
However, recent work pointed out that simple homogeneous graph model without meta-path can also achieve comparable results, which calls into question the necessity of meta-path. 
In this paper, we first present the intrinsic difference about meta-path-based and meta-path-free models, i.e., how to select neighbors for node aggregation.
Then, we propose a novel framework to utilize the rich type semantic information in heterogeneous graphs comprehensively, namely HAGNN (Hybrid Aggregation for Heterogeneous GNNs).
The core of HAGNN is to leverage the meta-path neighbors and the directly connected neighbors simultaneously for node aggregations.
HAGNN divides the overall aggregation process into two phases: meta-path-based intra-type aggregation and meta-path-free inter-type aggregation. 
During the intra-type aggregation phase, we propose a new data structure called fused meta-path graph and perform structural semantic aware aggregation on it.
Finally, we combine the embeddings generated by each phase.
Compared with existing heterogeneous GNN models, HAGNN can take full advantage of the heterogeneity in heterogeneous graphs. 
Extensive experimental results on node classification, node clustering, and link prediction tasks show that \mymodel{} outperforms the existing modes, demonstrating the effectiveness and efficiency of \mymodel{}.
\end{abstract}

\begin{IEEEkeywords}
Heterogeneous graph, Graph neural network, Graph representation learning, Hybrid node aggregation.
\end{IEEEkeywords}

\section{Introduction}
\IEEEPARstart{M}{any} real-world data can be naturally represented as graph structure. 
Meanwhile, in many practical scenarios, such as knowledge graphs~\cite{ji2021survey}, scholar networks~\cite{sen2008collective, horawalavithana2022expert, qian2021distilling}, and biochemical networks~\cite{zitnik2017predicting}, the graphs are heterogeneous. 
Compared to homogeneous graphs, heterogeneous graphs have more than one type of nodes or links, which encode more semantic information~\cite{sun2012mining}.

\begin{figure}
	\centering
	\includegraphics[width=1\columnwidth]{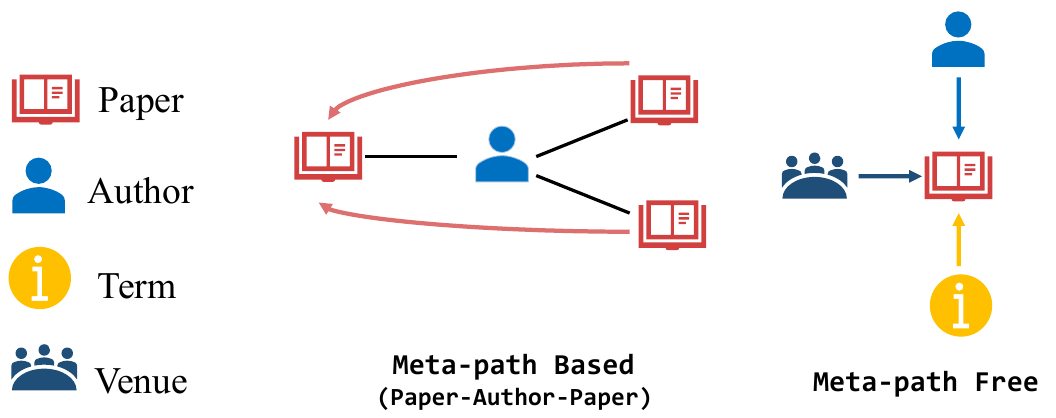}
	\vspace{-3ex}
	\caption
	{
	  Different aggregation schemes on the DBLP dataset.
	}
	\label{fig:f1}
	\vspace{-3ex}
\end{figure}

Graph neural networks (GNNs)~\cite{wu2020comprehensive, hamilton2017inductive} have achieved remarkable success in graph-structured data learning a low-dimensional representation for each node. 
Moreover, to tackle the challenge of heterogeneity, many representation learning models are proposed to utilize the rich semantic information in heterogeneous graphs. 
Among these methods, meta-path~\cite{sun2012mining} is considered a natural way to decouple diversified connection patterns between nodes. 
Specifically, meta-path is a composite relation consisting of multiple edge types.
For example, in Figure~\ref{fig:f1}, Paper-Author-Paper is a typical meta-path, which reflects that two papers are published by the same author.
metapath2vec~\cite{dong2017metapath2vec} formalizes meta-path-based random walks to compute node embeddings.
HAN~\cite{wang2019heterogeneous} and MAGNN~\cite{fu2020magnn} employ hierarchical attention to aggregate information from meta-path-based neighbors. 
GTN~\cite{yun2019graph} implicitly learns meta-paths by combining different node types based on the attention mechanism. 
Meta-path is also used for knowledge distillation~\cite{wang2022collaborative}, text summarization~\cite{song2022hierarchical}, and contrastive learning~\cite{wang2021self, jin2022heterogeneous} on heterogeneous graphs. 

\begin{table*}
  \caption{Meta-path related statistics on the DBLP and ACM datasets.}
  \centering
  \resizebox{0.8\textwidth}{!}{
  \label{Meta-path related statistics on DBLP dataset}
  
  \begin{tabular}{cccccc}
    \toprule
    Datasets &\#Nodes of types & \makecell{Type-specific  \\meta-path}  & \makecell{\#Edges in the\\ meta-path-based graph}  & \makecell{Average \\degree} & Information redundancy \\
    \midrule
    \multirow{3}{*}{DBLP}&\multirow{3}{*}{A (Author): 4057} &APA & 11113 & 3 & \multirow{2}{*}{APA-APTPA: 100\%} \\
    \multirow{3}{*}{}&\multirow{3}{*}{} & APTPA & 5000495& 1232& \multirow{2}{*}{APTPA-APCPA: 61.84\%} \\
    \multirow{3}{*}{}& \multirow{3}{*}{} &APCPA & 7043572 & 1736 & \multirow{2}{*}{} \\
    \midrule
    \multirow{2}{*}{ACM}&\multirow{2}{*}{P (Paper): 3025} &PAP  & 29767  & 10 & \multirow{2}{*}{PAP-PSP: 62.14\%} \\
    \multirow{2}{*}{}&\multirow{2}{*}{}  &PSP & 18499& 6& \multirow{2}{*}{} \\
    \bottomrule
  \end{tabular}
  }
\end{table*}
Meta-path plays an essential role in existing heterogeneous GNNs.
However, recent researchers~\cite{lv2021we} experimentally found that homogeneous GNNs such as GAT~\cite{velivckovic2017graph} actually perform pretty well on heterogeneous graphs by revisiting the model design, data preprocessing, and experimental settings of the heterogeneous GNNs, which calls into question the necessity of meta-paths~\cite{lv2021we,zhao2022space4hgnn}.
To answer this question, we first give an in-depth analysis about the intrinsic difference about meta-path-based models (e.g., HAN, MAGNN, GTN) and meta-path-free models (e.g., RGCN~\cite{schlichtkrull2018modeling}, GAT, SimpleHGN~\cite{lv2021we}). 
For a given node, how to select neighbors for node aggregation is a fundamental step in heterogeneous GNNs, which is also the key difference between the two types of models. 
Meta-path-based models usually construct the meta-path-based graph for the target node type. 
In the meta-path-based graph, all nodes have the same type and any two adjacent nodes have at least one path instance following the specific symmetric meta-path. 
The final representation of a node is calculated by aggregating its neighbors in the meta-path-based graph. 
For meta-path-free models, they directly aggregate neighbors in the original graph. 
In most heterogeneous graphs, each node and its immediate neighbors have different node types.
%
Thus, the final embedding is the aggregation of the nodes with different types.

Take an example as shown in Figure~\ref{fig:f1}, if we want to know the category of a paper, meta-path-based models check the categories of other papers written by the same author, while meta-path-free models collect information about the author of the paper, the conference where the paper was published, and the term of the paper etc. 
Overall, the immediate neighbors of a node contain key attributes, and the meta-path-based neighbors can easily supply high-order connectivity information of the same node type. 
Both meta-path-based neighbors and immediate neighbors are useful, they can complement each other.
Therefore, to improve the performance of heterogeneous GNNs, it is essential to design a new representation leaning method that can leverage both meta-path-based neighbors and immediate neighbors.

\begin{figure}[t!]
  \centering
      \subfigure[DBLP]{
    \includegraphics[width=0.32\columnwidth]{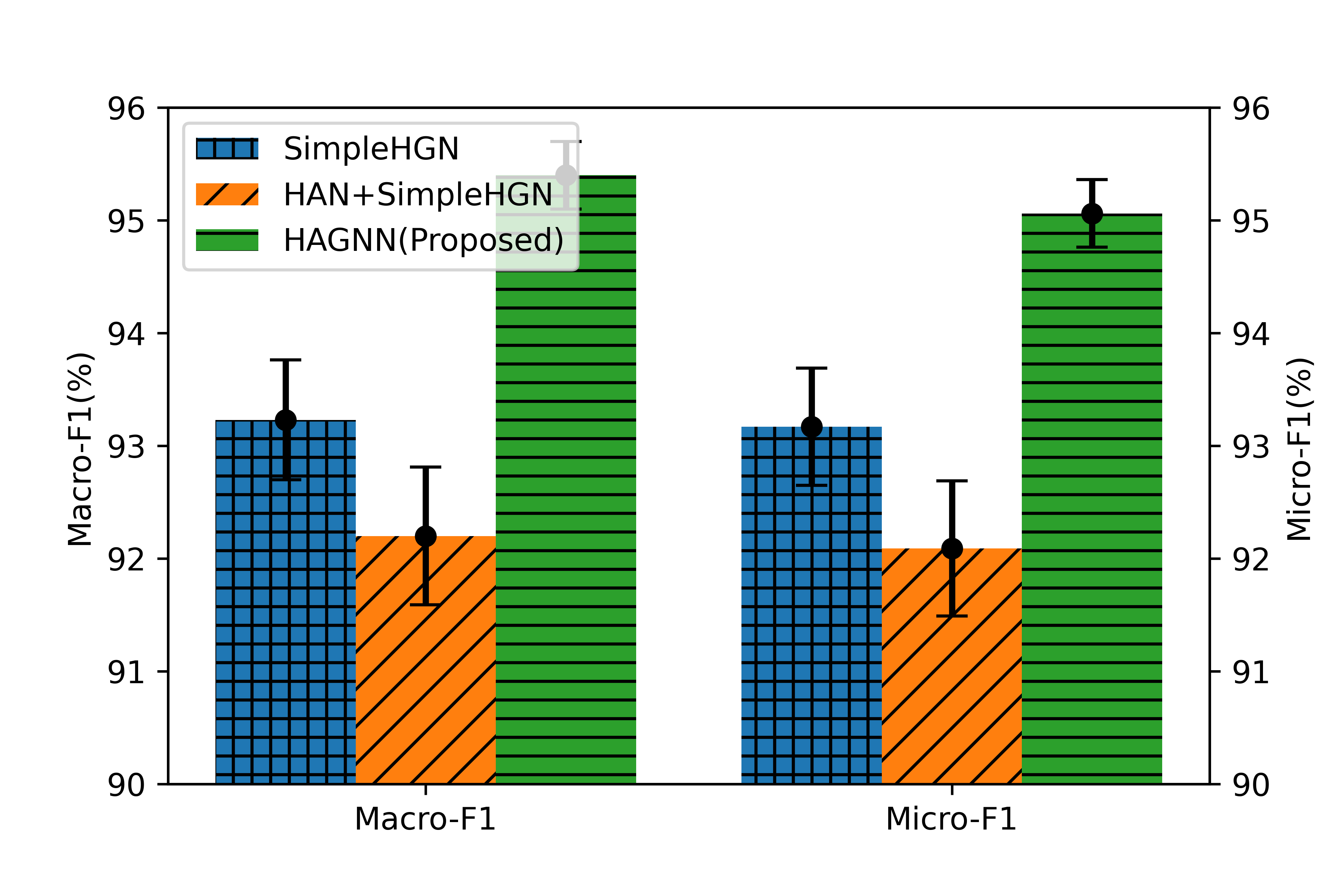}
    }
    \hspace{-3ex}
    \subfigure[IMDB]{
    \includegraphics[width=0.32\columnwidth]{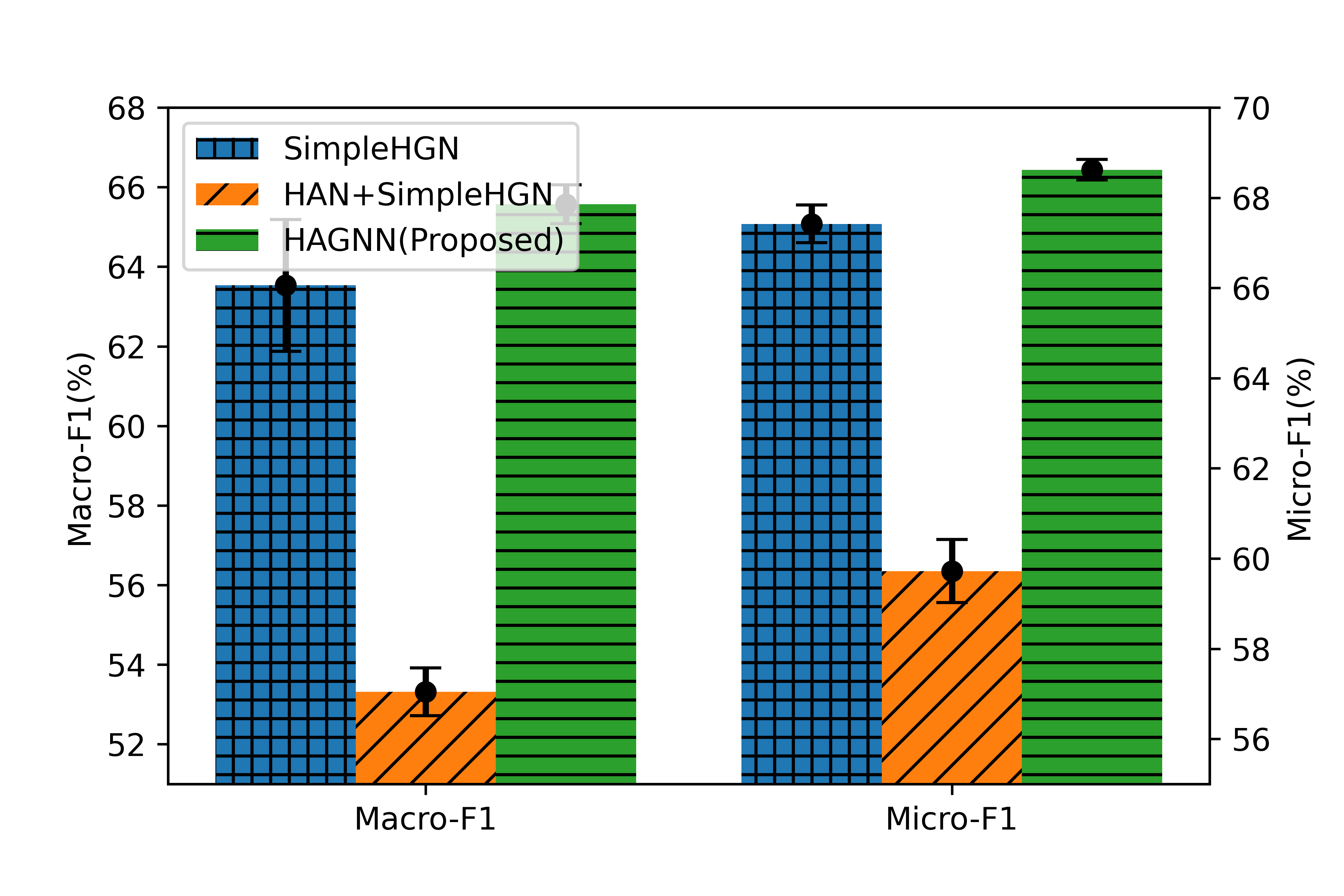}
    }
    \hspace{-3ex}
    \subfigure[PubMed]{
    \includegraphics[width=0.32\columnwidth]{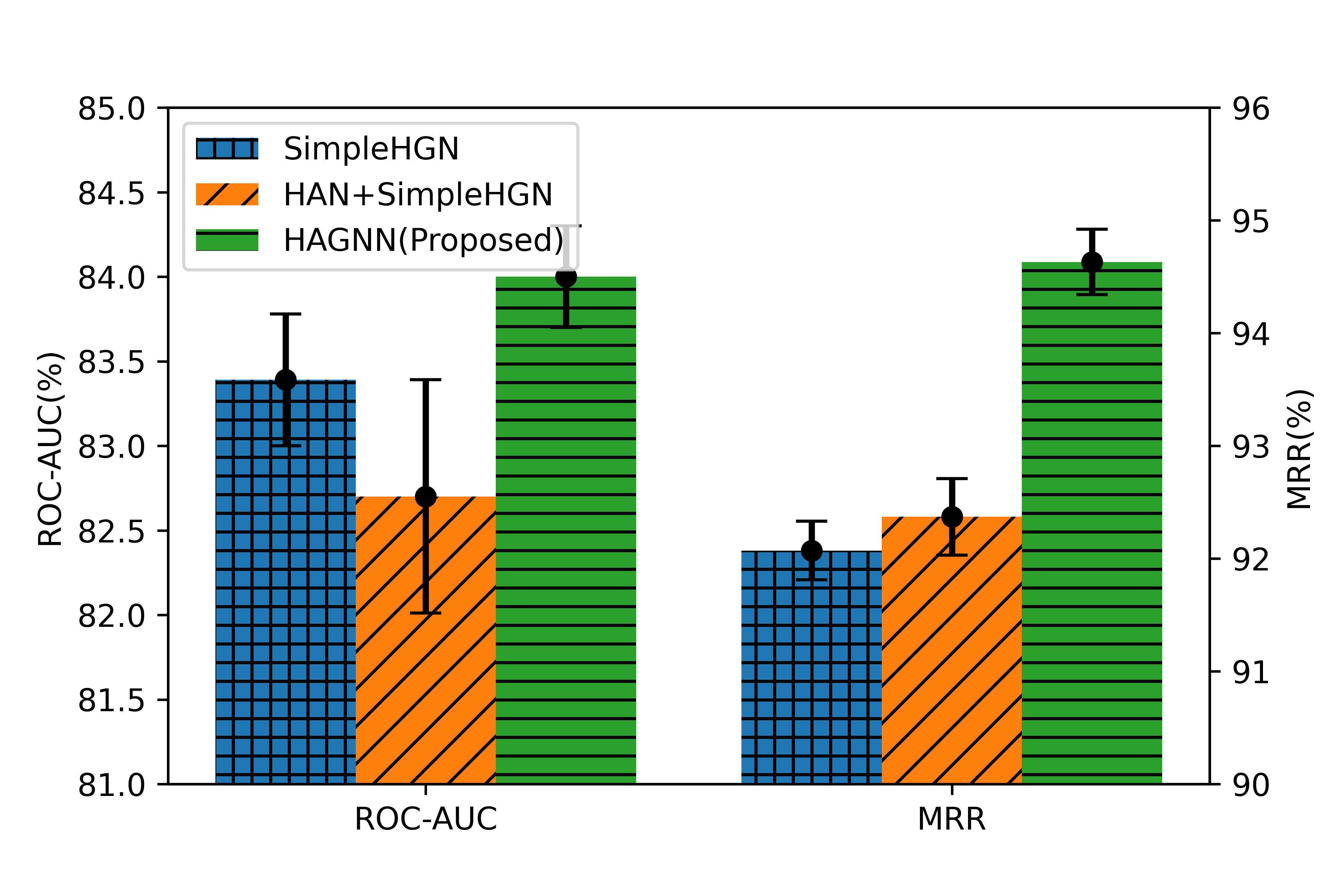}
    }
  \vspace{-1ex}
  \caption{Comparison between the simple combination of intra-type aggregation and inter-type aggregation (i.e., HAN + SimpleHGN) vs. \mymodel{} (Proposed)}
  \label{fig:han-simplehgn}
  \vspace{-3ex}
\end{figure}

A straightforward idea is to directly combine the meta-path-based intra-type aggregation with the immediate-neighbor-based aggregation.
But as shown in Figure \ref{fig:han-simplehgn}, simply combining the typical meta-path-based model (i.e., HAN) with the SOTA meta-path-free model (i.e., SimpleHGN) even leads to performance decreases on both the node classification datasets (i.e., DBLP and IMDB) and the link prediction dataset (i.e., PubMed), especially on the IMDB dataset.
%
The reason for this problem is that existing meta-path-based models suffer from information redundancy and excessive additional parameters, which may lead to over-parameterization as well as over-fitting if directly further combined with meta-path-free models.

Table~\ref{Meta-path related statistics on DBLP dataset} shows the issue of information redundancy, the definition of which can be seen in Definition 5 (Section~\ref{s:preliminary}). 
In existing meta-path-based models~\cite{wang2019heterogeneous,fu2020magnn}, each meta-path corresponds to a graph.  
The node representations learned from each separate math-path-based graphs are then aggregated with hierarchical attention.
For the DBLP dataset in Table~\ref{Meta-path related statistics on DBLP dataset}, the meta-path-based-graph produced by Author-Paper-Author (APA) is a subgraph of that produced by Author-Paper-Conference-Paper-Author (APCPA).
The meta-path-based graph generated by the meta-path Author-Paper-Conference-Paper-Author and  the meta-path Author-Paper-Term-Paper-Author (APTPA) have 61.84\% duplicate edges.
There exists the similar problem for the ACM dataset.
%
It can be seen that if we build separate meta-path-based graphs for each different meta-paths, unnecessary computation will be squandered on redundant information. 
Moreover, too many duplicate edges lead to excessive additional learnable parameters for node aggregation, which may degrade the learning performance.

Furthermore, Table~\ref{Meta-path related statistics on DBLP dataset} also shows that the number of meta-path-based neighbors is much larger than that of direct neighbors in the original graph and thus too many neighbors in the meta-path-based graph cause difficulties in the learning of attention weights. 
Existing meta-path-based models only consider the node connectivity in the meta-path-based graph, ignoring the \emph{structural semantic information} (e.g., the number of path instances following the specific meta-path), which can be exploited to improve the learning of node representation. 

Based on the above analysis, we propose a novel framework to utilize the rich type semantic information in heterogeneous graphs comprehensively, namely \mymodel{}\footnote{HAGNN is available at https://github.com/PasaLab/HAGNN} (Hybrid Aggregation for Heterogeneous Graph Neural Networks). 
The core of \mymodel{} is to leverage the meta-path-based neighbors and the directly connected neighbors simultaneously for node aggregation. 
Specifically, we divide the overall aggregation process into two phases: meta-path-based intra-type aggregation phase and meta-path-free inter-type aggregation phase.
During the intra-type aggregation phase, we first propose a new data structure called \emph{fused meta-path graph} to avoid information redundancy. 
For a specific node type, the meta-path neighbor relationships of multiple meta-paths are fused in a single graph, where all nodes have the same type. 
Also, the fused meta-path graph contains the connectivity information of multiple meta-graphs. %
Then, we perform attention-based intra-type aggregation in the fused meta-path graph. 

To further improve the learning of attention weights, we propose a \emph{structural semantic aware aggregation} method.
For each two neighbor nodes of the fused meta-path graph, we view the number of path instances in the original graph as the structural semantic weight, which is used to guide the learning of attention weights.
During the inter-type aggregation phase, we directly perform node aggregation with the self-attention mechanism in the original heterogeneous graph to capture the information of immediate neighbors. 
Finally, the node embeddings generated by the intra-type aggregation and the inter-type aggregation are combined.

To summarize, the main contributions are highlighted as follows:

\begin{itemize}
    \item \textbf{Novel hybrid aggregation mechanism.} Based on the analysis that both meta-path-based and meta-path-free aggregation should be beneficial to the heterogeneous graph, we propose a novel hybrid aggregation mechanism consisting of three stages: meta-path-based intra-type aggregation, meta-path-free inter-type aggregation, and combination of semantic information at different stages.

    \item \textbf{Novel data structure for meta-path-based aggregation.} To eliminate information redundancy and make the intra-type aggregation more effectively, we propose a simple but effective data structure called fused meta-path graph, which can efficiently capture the meta-path neighbors for a specific node type.
    
    \item \textbf{Structural semantic aware aggregation.} To improve the leaning of attention weights in the intra-type aggregation phase, we propose a structural semantic aware aggregation mechanism, which leverages the number of path instances as the auxiliary aggregation weights.
    
    \item \textbf{Effectiveness and efficiency.} Extensive experimental results on five heterogeneous graph datasets reveal that \mymodel{} outperforms the existing heterogeneous GNNs in terms of effectiveness and efficiency on node classification, node clustering, and link prediction tasks. The discussion about \mymodel{} also provides insightful guidance to the use of meta-paths in the heterogeneous graph neural networks.
\end{itemize}

\section{RELATED WORK}
\subsection{Homogeneous Graph Representation Learning}
Graph representation learning aims to learn low-dimensional representations from non-Euclidean graph structure. 
For homogeneous graphs, most methods learn node representations from neighborhood. Line~\cite{tang2015line} utilizes the first-order and second-order proximity between nodes to learn node embeddings.
DeepWalk~\cite{perozzi2014deepwalk}, node2vec~\cite{grover2016node2vec}, TADW~\cite{yang2015network}, and Struc2vec~\cite{ribeiro2017struc2vec} extract a node sequence by random walk and feed the sequence to a skip-gram model. 
Graph neural networks, following the message passing framework, have been widely exploited in graph representation learning. 
GNNs can be divided into spectral-based and spatial-based models~\cite{wu2020comprehensive}. 
GCN~\cite{kipf2016semi} is a typical spectral-based model that achieves spectral graph convolution via localized first-order approximation. 
While the spatial-based models such as GAT~\cite{velivckovic2017graph} leverage the attention mechanism for node aggregation.
Moreover, real-world graphs are often noisy and contain task-irrelevant edges.
To improve the generalization performance of GNNs by learning to drop task-irrelevant edges, robust graph neural networks via topological denoising~\cite{10.1145/3437963.3441734} or neural sparsification~\cite{10.5555/3524938.3526000} have been proposed.
To explain the predictions of a set of instances, PGExplainer~\cite{NEURIPS2020_e37b08dd} introduces a parameterized explainer for GNNs. 
To improve the robustness of GNNs, an efficient graph attack method that selects the vulnerable nodes as attack targets has been proposed~\cite{PGA}.
Inspired by the success of neural architecture search (NAS)~\cite{OLES}, PSP~\cite{PSP} employs progressive space pruning for graph NAS.

Representation learning models for homogeneous graphs are considered unsuitable for heterogeneous graphs due to ignoring type information, but recent work points out that some homogeneous graph models actually perform well in heterogeneous graphs, which is thought-provoking.

\subsection{Heterogeneous Graph Representation Learning}
Heterogeneous graphs have more than one type of nodes or edges. 
To utilize the semantics encoded in heterogeneous graphs, many models designed for heterogeneous graphs are proposed~\cite{wang2019heterogeneous, wang2022survey, AutoAC, CasMLN}. 

Depending on whether using meta-path, we can divide these models into meta-path-based and meta-path-free models.
For meta-path-based models, metapath2vec~\cite{dong2017metapath2vec} utilizes the node paths traversed by meta-path-guided random walks to model the context of a node. 
HIN2Vec~\cite{fu2017hin2vec} carries out multiple prediction training tasks to learn latent vectors of nodes and meta-paths.
HAN~\cite{wang2019heterogeneous} leverages the semantic information of meta-paths, and uses hierarchical attention to aggregate neighboring 
nodes.
MAGNN~\cite{wang2019heterogeneous} utilizes RotatE~\cite{sun2019rotate} to encode intermediate nodes along each meta-path and mix multiple meta-paths using hierarchical attention. 
GTN~\cite{yun2019graph} learns a soft selection of edge types and composite relations for generating useful meta-paths.
SHGNN~\cite{xu2021shgnn} uses a tree-based attention module to aggregate information on the meta-path and consider the graph structure in multiple meta-path instances.
R-HGNN~\cite{yu2022heterogeneous} learns node representations on heterogeneous graphs at a fine-grained level by considering relation-aware characteristics.
CKD~\cite{wang2022collaborative} learns the meta-path-based embeddings by collaboratively distilling the knowledge from intra-meta-path and inter-meta-path simultaneously.

The meta-path-free models extract rich semantic information without meta-path.
RGCN~\cite{schlichtkrull2018modeling} introduces relation-specific transformations to handle different edge types. 
HetGNN~\cite{zhang2019heterogeneous} uses Bi-LSTM to aggregate node features for each type and among types. 
SimpleHGN~\cite{lv2021we} revisits existing models and proposes a simple framework with GAT as backbone. 
Haar-MGL~\cite{Haar-MGL} proposes a framework that combines multimodal data, including visual, textual, and acoustic modalities that reflect the students’ personalities, their demographic information, their learning behavior and attention, with graph learning techniques. 
3D Haar semi-tight framelet transform is introduced to facilitate multimodal data fusion.
\cite{PEGFAN} provides a novel and general method to
construct Haar-type graph framelets having the permutation
equivariance property for heterophilous graph learning.
GCN-RW~\cite{GCN-RW} proposes a novel model termed graph convolutional networks with random weights by revising the convolutional layer with random filters and simultaneously adjusting the learning objective with regularized least squares loss.

\subsection{Difference with Existing Heterogeneous GNNs}
The meta-path-based heterogeneous GNN models essentially approximate the heterogeneous graph using multiple meta-path graphs.
Consequently, it requires the careful selection of appropriate meta-paths to minimize the margin between the meta-path graphs and the original heterogeneous graph. 
For instance, DiffMG~\cite{DiffMG} introduces an innovative approach by automatically generating meta-path graphs through trainable methods. It employs gradient descent in its methodology to iteratively minimize the margin. 
SeHGNN~\cite{SeHGNN} abandons excessive parameters to average all meta-paths within $N$ hops, in order to minimize this margin with as many meta-paths as possible.
In contrast, meta-path-path free models assumes the existence of a function $F$, which can encode all semantic information in the heterogeneous graph. For example, HetGNN~\cite{zhang2019heterogeneous} employs Bi-LSTM, while SimpleHGN~\cite{lv2021we} utilizes edge type embeddings.

In this paper, we propose a completely new direction for designing heterogeneous GNNs. 
The focus is neither on reducing the approximation margin nor on creating a more powerful semantic extraction function. 
Instead, it designs a reasonable framework that allows these two aspects to work together in a synergistic manner, achieving complementarity.
Actually, the difference between meta-path-based and meta-path-free models is that they have conflict on which kind of neighbors is more informative. 
In this paper, we propose a hybrid aggregation mechanism that can leverage both the meta-path-based neighbors and immediate neighbors effectively.

\section{PRELIMINARIES AND NOTATIONS}
\label{s:preliminary}
\begin{table}[!t]
    \caption{Notations used in this paper.}
    \label{tab:notation}
    \begin{tabular}{l|l}
        \toprule
        \textbf{Notations} & \textbf{Definitions}\\
        $\mathbb{R}^{n}$ & $n$-dimensional Euclidean space\\
        $\mathbf{A}$ &  Adjacent matrix\\
        $\mathcal{G}$ & A homogeneous graph  \\
        $\mathcal{V}$ & The set of nodes in a graph\\
        $\mathcal{E}$ & The set of edges in a graph\\
        $\mathcal{H}$ & A heterogeneous graph  \\
        $\mathcal{T}$ & The set of node types in a heterogeneous graph\\
        $\mathcal{R}$ & The set of edge types in a heterogeneous graph\\
        $v$ & A node $v\in\mathcal{V}$\\
        $t$ & A node type $t\in\mathcal{T}$\\
         $r$ & An edge type $r\in\mathcal{R}$\\
        $p_t$ & A meta-path of type $t$\\
        $\mathcal{P}_t$ & The set of meta-paths of type $t$   \\
        $G_p$ & Meta-path $p$ based homogeneous graph (Definition 4) \\
        $\mathcal{N}_{G_p}^v$ &The set of neighbors of node $v$ in $G_p$\\
        $x_v$ & Raw feature (attribute) vector of node $v$\\
        $h_v$ & Hidden state (embedding) of node $v$\\
        $z_v$ & Output state (embedding) of node $v$\\
        $y_v$ & One-hot label vector of node $v$\\
        $W$ & Parameter matrix\\
        $\alpha, \delta$ & Attention weight\\
        $\sigma(\cdot)$ & Activation function\\
        $|\cdot|$ & The cardinality of a set\\
        $\|$ & Vector concatenation\\
        \bottomrule
    \end{tabular}
\end{table}

\begin{myDef}
\textbf{Heterogeneous Graph \cite{sun2012mining}}. A heterogeneous
graph is a directed graph with the form $\mathcal{H} = \{ \mathcal{V}, \mathcal{E}, 
 \mathcal{T}, \mathcal{R} , \mathcal{\phi}, 
  \mathcal{\psi}  \} $, where $\mathcal{V}$ and $\mathcal{E}$ denote the node set and the edge set in $\mathcal{H}$. Each node $v_i \in \mathcal{V}$ is  associated with a node type $\phi(v_i) = t_i \in \mathcal{T}$. Similarly, each edge $e_{ij} \in \mathcal{E}$ is associated with an edge type $\psi(e_{ij}) = r_{ij} \in \mathcal{R}$. 
  %
  In graph $\mathcal{H}$, $|\mathcal{T}| + |\mathcal{R}| > 2$. 
  Every node has attribute $x_i \in \mathbb{R}^{d_i} $, and $d_i$ varies from the type of node $v_i$.
  Let $\mathbf{A}_{t_i,t_j}$ denote the adjacent matrix of type $t_i$ and $t_j$. 
  $\mathbf{A}_{t_i,t_j}[u][v] = 1$ indicates that $\phi(u) = t_i$ , $\phi(v) = t_j$ and nodes $u$ ,$v$ are connected.
\end{myDef}

 \begin{myDef}
\textbf{Meta-Path~\cite{sun2012mining}}. A meta-path $p$ is a composite relation, which consists of multiple edge types, i.e., $p = t_1 \xrightarrow{r_1} t_2 \xrightarrow{r_2} \dots \xrightarrow{r_l} t_{l+1}$, where $t_1,\dots,t_{l+1} \in \mathcal{T}$ and $r_1,\dots,r_{l} \in \mathcal{R}$. One meta-path contains many meta-path instances in $\mathcal{H}$. In this paper, we use meta-pathes satisfying $t_1 = t_l+1$
\end{myDef}
 \par
 \begin{myDef}
\textbf{Meta-path-based Neighbors\cite{fu2020magnn}}. Given a meta-path $p$ in $\mathcal{H}$, the meta-path-based neighbors of node $v$ is defined as the set of nodes that connect with node $v$ via a meta-path instance of $p$. If meta-path $p$ is symmetrical, the meta-path-based neighbors of node $v$ contain itself.
 \end{myDef}
 \par
 
 \begin{myDef}
\textbf{Meta-path-based Graph\cite{fu2020magnn}}. Given a meta-path $p$ in $\mathcal{H}$, the meta-path--based graph $G_p$ of $p$ is a graph
constructed by all meta-path-based neighbor pairs.
$G_p$ is homogeneous if the head and tail node types of $p$ are the same. 
The neighbors of node $v$ in $G_p$ can be donated as $N^v_{G_p}$.
 \end{myDef}

\begin{myDef}
{
\textbf{Information redundancy of meta-path-based graphs}. Information redundancy is the ratio of duplicated edges. Given two meta-path $p_1$, $p_2$ and corresponding meta-path-based graphs $G_{p_1}$, $G_{p_2}$, the information redundancy (IR) between the two graphs is $$
IR (G_{p_1},G_{p_2}) = \frac{EgdeSet(G_{p_1}) \cap EgdeSet(G_{p_2})}{EgdeSet(G_{p_1}) \cup EgdeSet(G_{p_2})}
$$
}
\end{myDef}
 
Table~\ref{tab:notation} shows the used notations and their definitions.

\section{THE PROPOSED METHODOLOGY}
%
\subsection{Overall Framework}
\begin{figure*}
	\centering
	\includegraphics[width=1\textwidth]{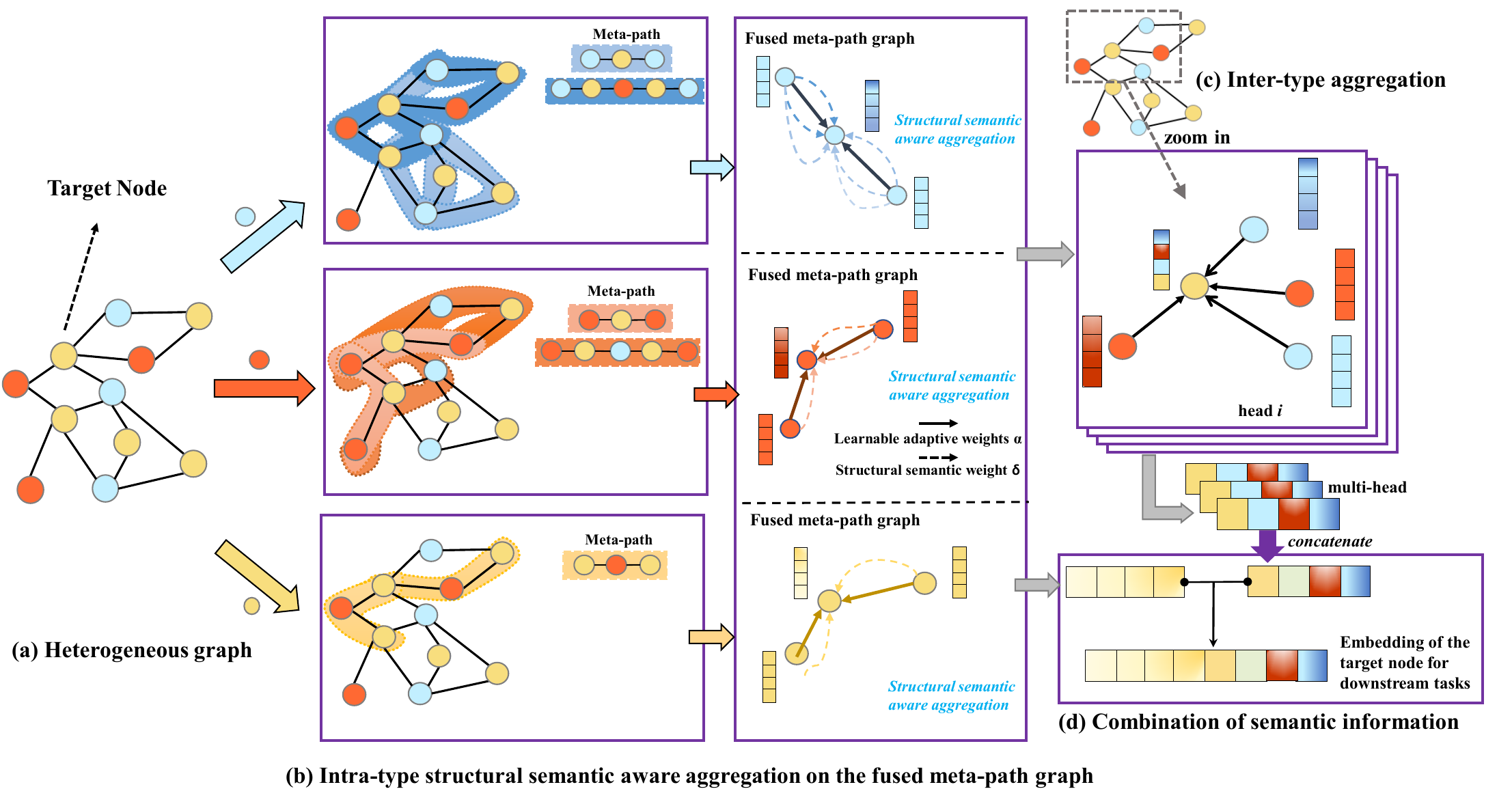}
	\vspace{-4ex}
	\caption
	{
		The overall framework of \mymodel{}. Different circle colors represent different node types.
	}
	\label{fig:framework}
	\vspace{-1ex}
\end{figure*}
Figure~\ref{fig:framework} shows the overall framework of \mymodel{}, which consists of three phases: meta-path-based intra-type aggregation, meta-path-free inter-type aggregation, and combination of semantic information.
In the intra-type aggregation phase, we first construct the fused meta-path graph that contains all meta-path-based neighbors.
%
Then, we leverage the structural semantic weights to guide the node aggregation in the fused meta-path graph.
The node embeddings obtained in the intra-type aggregation phase are fed to the inter-type aggregation phase, which performs node aggregation directly on the original heterogeneous graph.
The embeddings generated by the two aggregation phases are combined to form the final embeddings of the target node types, which are further used for downstream tasks.
%

%
The intra-type aggregation phase aims to capture the information of high-order meta-path neighbors, while the inter-type aggregation phase directly aggregates the attribute information of immediate neighbors.
The two phases perform node aggregation from two different perspectives.
%
Next, we introduce each phase of \mymodel{}.

\subsection{Meta-path-based Intra-type Aggregation}
Since the intra-type aggregation is performed on the fused meta-path graph, we first introduce the definition of the proposed fused meta-path graph.
\subsubsection{Fused Meta-path Graph}
The core of heterogeneous GNNs is to use the graph topology to perform message aggregation.
Thus, the graph topology plays an essential role in heterogeneous GNNs.
Most of existing heterogeneous GNNs construct the meta-path-based graph for node aggregation. 
Each meta-path corresponds to a meta-path-based graph.
%
As shown in Table~\ref{Meta-path related statistics on DBLP dataset}, information redundancy arises when we put meta-path-based neighbors of different meta-paths in separate graphs.
A large number of duplicate edges leads to computational redundancy and excessive additional parameters when computing and optimizing attention weights in the meta-path-based graph, leading to negative impacts on the computation efficiency and learning performance.
To address the issues, we propose a novel data structure called fused meta-path graph to carry the information of multiple meta-paths.
\begin{table}[!t]
    \caption{Different methods to generate homogeneous graphs using meta-paths ( Detailed experimental results on the DBLP dataset can be referred to Section \uppercase\expandafter{\romannumeral 5}.)}
    \label{tab:differentway}
    \resizebox{\columnwidth}{!}{
    \begin{tabular}{c|c|c|c}
        \toprule
        \multicolumn{4}{c}{ \thead{Given the meta-path set of type $t$,  $ 
        \mathcal{P}_t =\{p_{t}^1,p_{t}^2,\dots,,p_{t}^{l} \}$ \\  for each $p_t^i \in \mathcal{P}_t$, $p_t^i = t_1^i \xrightarrow{r_1^i} t_2^i \xrightarrow{r_2^i} \dots \xrightarrow{r_l^i} t_{l+1}^i $ }} \\
        Models & Methods to extract homogeneous graphs & Time Overhead & Performance \\
        \midrule
        HAN \cite{wang2019heterogeneous}, MAGNN\cite{fu2020magnn} & $\thead{\begin{cases}
            h_1 = f_{GAT}(G_{p^1_{t}})\\
             h_2 = f_{GAT}(G_{p^2_{t}}),\\
             \dots,\\
             h_{|\mathcal{P}_t|} = f_{GAT}(G_{p^{l}_{t}}),\\
        \end{cases} \\
        \\
        z = ATTENTION (h_1,h_2,\dots,h_{|\mathcal{P}_t|}) \\
        }$ & \makecell{medium \\ (HAN 0.23s/epoch\\
        MAGNN 19s/epoch)} & medium \\
        
        \midrule
        
         GTN \cite{yun2019graph} & \thead{
         $ \mathbb{A}_i = \{\mathbf{A}_{r_1^i}, \mathbf{A}_{r_2^i}, \dots, \mathbf{A}_{r_l^i}\} $ \\
         $ \mathbb{Q}_{i} = \phi(softmax(W^{\phi}_i),\mathbb{A}_i) $ \\
         $ G_{p_t^i} = \prod_{j=1}^{l} \mathbf{Q}_{r^i_j} $ \\
         where $\phi$ is 1 $\times$ 1 convolution \\ 
         $ z = f_{GCN}(G_{p_t^i}) $ \\
         } & \makecell{very high \\ (340s/epoch)} & medium\\
         
         \midrule
         
        \textbf{HAGNN (ours)} &  \textbf{\thead{ $ \mathcal{G}_t = \bigcup_{p_{t} \in \mathcal{P}_t} G_{p_t} $ \\
        $ z = f_{HAGNN}(\mathcal{G}_t,\mathcal{H}) $ \\ }
        } & \textbf{\makecell{low \\ (0.17s/epoch)}} & \textbf{high} \\
        \bottomrule
    \end{tabular}}
\end{table}

\begin{myDef}
\textbf{Type of meta-path:} Given a meta-path $p$ with length of $l$, $p = t_1 \xrightarrow{r_1} t_2 \xrightarrow{r_2} \dots \xrightarrow{r_l} t_{l+1}$, where $t_1,\dots,t_{l+1} \in \mathcal{T}$ and $r_1,\dots,r_{l} \in \mathcal{R}$. The type of $p$ is the head node type $t_1$ and meta-path $p$ can also be denote as $p_{t_1}$.
\end{myDef}

 \begin{myDef}
\textbf{Fused meta-path graph:} Given a node type $t \in \mathcal{T}$ of a heterogeneous graph $\mathcal{H}$, the meta-path set of type $t$ is $ \mathcal{P}_t =\{p_{t}^1,p_{t}^2,\dots,p_{t}^{l}\}$, the fused meta-path graph $ \mathcal{G}_t$ is a single graph contains all meta-path neighbors in $\mathcal{P}_t$. 
The neighbors of node $v$ in $\mathcal{G}_t$ is denoted as $N_{\mathcal{G}_t}^{v}$.
\end{myDef}

$\mathcal{G}_t $ is a homogeneous graph and defines the topological adjacency relationship between nodes of the same node type. 
To construct $\mathcal{G}_t$, we first obtain the meta-path-based graph $G_{p_t}$ for $p_{t} \in \mathcal{P}_t$, and then we take the union of $G_{p_t}$, i.e.,
 \begin{equation}
    \begin{gathered}
       \mathcal{G}_t = \bigcup_{p_{t} \in \mathcal{P}_t} G_{p_t}
    \end{gathered}
    \label{eq_g_union}
\end{equation}

Figure~\ref{fig:framework} shows the construction process of the fused meta-path graph. 
Unlike the meta-path-based graph, each node type instead of each meta-path corresponds to a fused meta-path graph. 
%
For node $v$ in fused meta-path graph $\mathcal{G}_t$, $N_{\mathcal{G}_t}^{v}$ contains all meta-path neighbors from $\mathcal{P}_t$.
The relationship between $v$ and its neighbor $u \in N_{\mathcal{G}_t}^{v}$ may belong to multiple meta-paths. 
Since the repeated meta-path-based neighbor relationships are reflected in a single edge, the information redundancy between different meta-paths can be eliminated.

Table~\ref{tab:differentway} compares fused meta-path graph with existing methods to extract homogeneous graphs from heterogeneous graphs using meta-paths.
For HAN and MAGNN, meta-path-based graphs are directly constructed by meta-path-based neighbors. 
But as we point out in Table~\ref{Meta-path related statistics on DBLP dataset}, there is a serious data redundancy problem between different meta-path graphs.
%
GTN is proposed to learn meta-path graphs. It uses 1x1 convolution to softly select different edge types, and then uses matrix multiplication to generate new meta-paths. 
Since the selection of edge types is continuous rather than discrete, the meta-path graph generated by GTN is dense. Moreover, the computation overhead of GTN is huge.
Overall, the proposed fused meta-path graph can solve the data redundancy problem and achieve better performance with less time overhead.
\begin{algorithm}[t]
\SetAlgoLined
\KwIn{Heterogeneous graph $\mathcal{H}$, \\
 \quad\quad\quad Meta-paths $\mathcal{P}$, \\ \quad\quad\quad Type selection $therehold$}
\KwOut{A set of fused meta-path graph $\mathcal{S}$}
\tcc{Type selection for intra-type aggregation}
$\mathcal{T}' = \{t \ | \  \frac{|\mathcal{V}_t|}{| \mathcal{V}|}  > threshold \text{ and $t$ has closed meta-paths} \}$
\tcc{Build the fused meta-path graph set}
$\mathcal{S}$ $\leftarrow \emptyset;$ \\
\For{$t \in \mathcal{T}' $}{
$\mathcal{G}_t$ $\leftarrow \emptyset;$ \\
    \For{$ p_t \in \mathcal{P}_t $}{
    construct meta-path based graph $G_{p_t}$; \\
    \quad $\mathcal{G}_t = \mathcal{G}_t \cup G_{p_t}$ \;
    }
    $\mathcal{S} = \mathcal{S} \cup \mathcal{G}_t$
}

\Return $\mathcal{S}$
 \caption{Construction of fused meta-path graph }
 \label{alg}
\end{algorithm}
\subsubsection{Type Selection for Intra-type Aggregation}
\mymodel{} is a two-phase aggregation model.
As shown in Figure~\ref{fig:framework}, in the latter phase (i.e, the inter-type aggregation phase), the representations of other node types are absorbed in the embeddings of the target node type.
%
%
Unlike the previous models~\cite{wang2019heterogeneous,fu2020magnn} where other node types only play a role as bridges between nodes of the target type and do not directly participate in the learning of node representation, we also perform intra-type aggregation for non-target node types.


Moreover, not all node types are suitable for participating in the intra-type aggregation phase.
We select node types for intra-type aggregation mainly based on the following two aspects.
First, in heterogeneous graphs, the quantity of different types of nodes is varied. 
For example, the DBLP dataset has 14,328 paper nodes, but only 20 conferences nodes. 
For those types with a small number of nodes, we believe that the relationship between the nodes is clear enough and the intra-type aggregation is not necessary.  
Second, having a closed meta-path is also an important condition. %
A meta-path $p$ is closed if its head and tail types are the same, and only using a closed meta-path can generate homogeneous graphs.

Formally, we denote the type set participating in the intra-type aggregation phase as $\mathcal{T}'$.
%
\begin{equation}
    \begin{gathered}
\begin{aligned}
\mathcal{T}' = \{t \ | \  \frac{|\mathcal{V}_t|}{| \mathcal{V}|}  > threshold \text{ and $t$ has closed meta-paths} \}
\end{aligned}
    \end{gathered}
    \label{eq_a_def}
\end{equation}
where $V_t$ denotes the number of nodes with type $t$.
In practice, the threshold can be set to 1\%.
Algorithm 1 describes the process of constructing the fused meta-path graph. 
First, only qualified node types are selected for intra-type aggregation.
Then, for each selected node type $t$, we construct the corresponding fused meta-path graph $\mathcal{G}_t$ by taking the union of $G_{p_t}$. 
Finally, we can get a set of fused meta-path graphs for all selected node types.

\begin{table}
  \caption{Comparison of the number of edges in meta-path-based graphs and fused meta-path graph}
  \resizebox{\columnwidth}{!}{
  \label{metapath comp}
  \centering
  \begin{tabular}{ccccc}
    \toprule
    Datasets   &Selected meta-path &\makecell{\#Edges in the\\ meta-path-based graphs}  & \makecell{\#Edges in the\\ fused meta-path graph} & Rate of reduction\\ 
    \midrule
     \multirow{3}{*}{DBLP} & APA &\multirow{3}{*}{} &\multirow{3}{*}{} &\multirow{3}{*}{}\\
    & APTPA & 12055180 & 7043572 & 41.57\%$\downarrow$ \\
     & APCPA&\multirow{3}{*}{} \\
     \midrule
    \multirow{2}{*}{FreeBase} & MBOM &\multirow{2}{*}{164482286} &\multirow{2}{*}{109146782} &\multirow{2}{*}{33.64\%$\downarrow$} \\
     & MBOBUM & & & \\
      \midrule
      \multirow{2}{*}{IMDB}& AMDMA &\multirow{2}{*}{172478} &\multirow{2}{*}{138272} &\multirow{2}{*}{19.83\%$\downarrow$} \\
      & AMA &  & &\\
    \bottomrule
  \end{tabular}
  }
  \vspace{-3ex}
\end{table}
%

Furthermore, we count the number of edges in the meta-path-based graphs and fused meta-path graph.
As Table~\ref{metapath comp} shows, the number of edges decreases by 41.57\%, 33.64\%, and 19.83\%, on DBLP, FreeBase, and IMDB, respectively. 
The experimental results demonstrate that the proposed fused meta-path graph can effectively reduce information redundancy, thus increasing training efficiency.

\subsubsection{Type-specific Linear Transformation}
Before the intra-type aggregation, since nodes of different types have different feature dimensions, we apply a type-specific linear transformation to each type of node, projecting the features of each type of node into the same latent factor space. For node $v \in \mathcal{V}$ of $t \in \mathcal{T}$:
\begin{equation}
    \begin{gathered}
\begin{aligned}
{\tilde h}_v = {\tilde W}_t \cdot {x}^{t}_{v}
\end{aligned}
    \end{gathered}
    \label{eq_liner}
\end{equation}
where $x^{t}_{v} \in \mathbb{R}^{d_t}$ is the original feature vector of node $v$.
The dimension $d_t$ varies from the node type.
${\tilde W}_t \in \mathbb{R}^{d \times d_t}$ is the learnable transformation matrix.

After the linear transformation, all nodes have the same dimension $d$. 
Next, the intra-type node aggregation can be carried out to aggregate meta-path-based higher-order neighbors.

\subsubsection{Structural Semantic Aware Aggregation}
For the specific node type $t$, once the corresponding fused meta-path graph $\mathcal{G}_t$ is constructed, we perform intra-type node aggregation according to the topology (i.e., the neighborhood of each node) of $\mathcal{G}_t$.
For each node $v$ in the fused meta-path graph, the neighborhood is much larger than that in the original graph.
Optimizing the attention weight of each neighbor becomes a challenging task.
To address the challenge, we propose a structural semantic aware mechanism to guide the learning of the attention weight for node aggregation. 

For meta-path $p_t$, suppose that the length of $p_t$ is $l$ and the node type sequence of $p_t$ is $t_1 \to t_2 
\to \dots \to t_{l-1} \to t_l $.
Let $\mathcal A$ be the weighted adjacent matrix following meta-path $p_t$:

\begin{equation}
    \begin{gathered}
       {\mathcal A}(\mathcal{H},p_t) = \prod_{i=1}^{l-1} \mathbf{A}_{t_i,t_{i+1}} 
    \end{gathered}
    \label{eq_n_cal}
\end{equation}
where $\mathbf{A}_{t_i,t_i+1} $ is the adjacent matrix of type $t_i$ and  $t_{i+1}$.
For the node pair $(u,v)$, ${\mathcal A}(\mathcal{H},p_t)[u][v]$ denotes the number of path instances from node $u$ to node $v$ in heterogeneous graph $\mathcal{H}$, and the node types on the path conform to the pattern of meta-path $p_t$. 
\begin{figure}
	\centering
	\includegraphics[width=0.95\columnwidth]{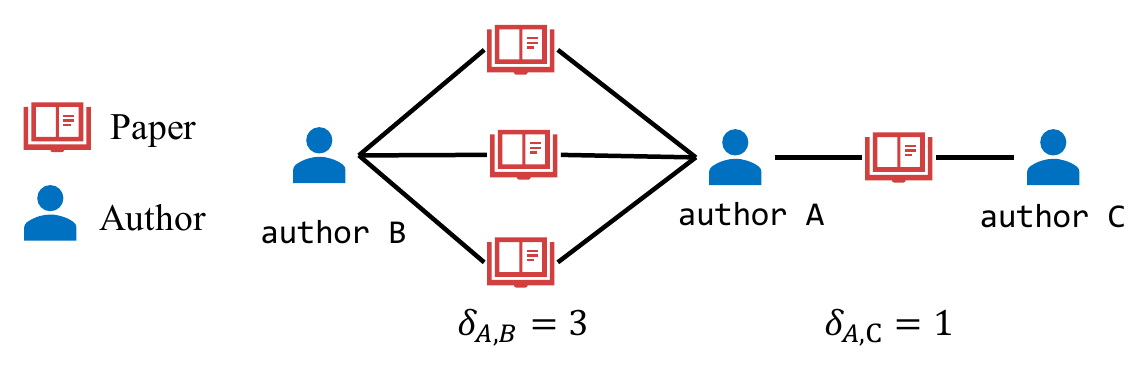}
	\caption
	{
		Structural semantic weight of the meta-path APA.
	}
	\label{fig:nvis}
	\vspace{-3ex}
\end{figure}

$\mathcal A$ defines the meta-path-based similarity between nodes of the same type. 
In the fused meta-path graph $\mathcal{G}_t$, since the neighbor pair $(u,v)$ may contain the meta-path neighbor relationships of multiple meta-paths, we further additively mix these meta-path-based similarities:

\begin{equation}
    \begin{gathered}
       \delta^{t}_{uv} = \sum_{p_t \in \mathcal{P}_t} {\mathcal A}(\mathcal{H},p_t)[u][v]
    \end{gathered}
    \label{eq_n}
\end{equation}
$\delta^{t}_{uv}$ can be viewed as the common neighbours~\cite{newman2001clustering, kossinets2006effects} of node $u$ and $v$ in $\mathcal{G}_t$. 
 We define $\delta$ as \textbf{\emph{structural semantic weight}} because it can reflect the structural semantic similarity between nodes.
 %
 For example, as shown in Figure~\ref{fig:nvis}, the number of papers published by author $A$ and author $B$ is more than the number of papers published by author $A$ and author $C$. 
 Thus, $A$ and $B$ should have a stronger relationship than $A$ and $C$, and their representations should be more similar. 

 To make structural semantic weight participate in node aggregation, we normalize it by softmax.
 
\begin{equation}
    \begin{gathered}
       {\tilde \delta}^{t}_{uv} = \frac{exp(\delta^{t}_{uv})}{\sum_{(u',v') \in {\mathcal{G}_t}} exp(\delta^{t}_{u'v'})}
    \end{gathered}
    \label{eq_n}
\end{equation}

However, only use $\delta$ as the attention weight for intra-type aggregation has the following two problems:
\begin{itemize}
\item $\delta$ is not capable of distinguishing the importance of different meta-paths by directly summing all meta-path-based similarity.
%
As the higher-order relationship between nodes, different meta-paths represent different levels of intimacy. 
%
For example, there are three nodes $A_1$, $A_2$, $A_3$, and the meta-path instances between them is $A_1-P_1-A_2$ and $A_1-P_1-T_1-P_2-A_3$. Even if the meta-path connected between A1-A2 and A1-A3 are different,  $\delta_{(A_1,A_2)} = \delta_{(A_1,A_3)}$ in the fused meta-path graph.
%
\item $\delta$ only reflects the intuitive semantic information.
As a fixed value, $\delta$ does not express small differences within the same meta-path. 
For example, for the APA meta-path, although co-authors may be in the same field, there are also papers in cross-cutting fields whose authors are in different fields.
\end{itemize}

Therefore, $\delta$ contains useful information, but cannot be completely relied upon. 
Learnable adaptive weights are also necessary. 
In this paper, we adopt a graph self-attention mechanism to calculate the adaptive weights.
\begin{equation}
    \begin{gathered}
        u \in N_{\mathcal{G}_t}^v \\
      e^{t}_{uv} = LeakyReLU(a_{t}^T \cdot W_t[{\tilde h_v} || {\tilde h_u}]) \\
      \alpha^{t}_{uv} = \frac{exp(e^t_{uv})}{\sum_{s \in N_{\mathcal{G}_t}^v} exp(e^t_{us}) }\\
    \end{gathered}
    \label{eq_alpha}
\end{equation}
%
where $||$ denotes the concatenation operator, $a_t$ and $W_t$ are learnable parameters in the self-attention mechanism that vary with node type $t$.
%
%
$\alpha$ represents learned attention weights, which can be adaptively adjusted according to the performance of downstream tasks.
Inspired by \cite{he2020realformer}\cite{lv2021we}, we introduce $\delta$ as an edge residual into $\alpha$:
\begin{equation}
    \begin{gathered}
      \eta^{t}_{uv} = (1-\beta)\alpha^{t}_{uv} + \beta {\tilde \delta}^{t}_{uv}
    \end{gathered}
    \label{eq_res}
\end{equation}
$\beta$ is a hyperparameter that controls how much structural semantic information we add into the attention weight. 
At each layer of intra-type aggregation, the attention weight between nodes perceives the structural semantic information in the fused meta-path graph, and the learnable parameters refine the structural semantic weight, helping the target node to select neighbors more effectively. 

Then, we can perform intra-type aggregation as follows:
\begin{equation}
h^{intra}_v = 
\begin{cases}
{\tilde h}_v& \phi(v) \notin \mathcal{T}' \\
\sigma (\sum_{u \in N^v_{\mathcal{G}_t} } (\eta^{t}_{uv} \cdot {\tilde h}_u)) & \phi(v) \in \mathcal{T}' \\
\end{cases}
    \label{eq_agg}
\end{equation}
For the node $v$ of type $t \in \mathcal{T}'$, the new embedding $h^{intra}_v$ is the weighted sum of its neighbors. 
Otherwise it stays the same. 
Then, after the intra-type aggregation phase, all type information of the heterogeneous graph is fully utilized, and then
$h^{intra}_v$ are fed to the inter-type aggregation phase.
\subsection{Meta-path-free Inter-type Aggregation}

The inter-type aggregation considers the direct neighbors $N^v_{\mathcal{H}}$ of node $u$ in the original graph $\mathcal{H}$. 
Since nodes and their first-order neighbors often belong to different types, the neighbors of a node reflect its attributes.
Thus, the inter-type aggregation is actually the process of continuously integrating the node attributes.
When performing information fusion between types, the contribution of different neighbors to the target node is different.
For each neighbor $u \in N^v_{\mathcal{H}}$, we can learn a normalized importance weight $\alpha_{uv}$.
\begin{equation}
    \begin{gathered}
      e_{uv} = LeakyReLU(a^{\mathrm{T}} \cdot W [h^{intra}_{v} || h^{intra}_{u}]),\\
      \alpha_{uv} = \frac{exp(e_{uv})}{\sum_{s \in N^v_{\mathcal{H}}} exp(e_{us}) } \\
      h^{inter}_{v} = \sum_{u \in N^v_{\mathcal{H}}} \alpha_{uv} \cdot h^{intra}_{u}
    \end{gathered}
    \label{eq_gat}
\end{equation}
where $a$ and $W$ are learnable parameters shared on all node types. 
Moreover, to stabilize the learning process and reduce the the large variation caused by the heterogeneity of $\mathcal{H}$, we further employ the multi-head attention mechanism.
%
Specifically, we implement $K$ independent attention processes and concatenate their outputs.
\begin{equation}
    \begin{gathered}
      h^{inter}_{v} = \Vert_{k=1}^{K} \sum_{u \in N^v_{\mathcal{H}}} \alpha_{uv}^k \cdot h^{intra}_{u}
    \end{gathered}
    \label{eq_mutilhead}
\end{equation}

\subsection{Combination of Semantic Information}
Due to the different neighborhoods selected in the aggregation phase, intra-type aggregation and inter-type aggregation actually extract the semantic information encoded in heterogeneous graphs from different perspectives.
To explicitly capture the information from these two perspectives, we further propose an information combination method.
\begin{equation}
    \begin{gathered}
      z_{v} = COMBINE(h^{intra}_{v} , W_m \cdot h^{inter}_{v})
    \end{gathered}
    \label{eq_fusion}
\end{equation}
%

First, we unify the dimensions of $h^{intra}_{v}$ and $h^{inter}_{v}$. $W_m \in \mathbb{R}^{ K d \times d}$ is the learnable transformation matrix.
Then, we combine the two embeddings by addition, concatenation or other pooling operations.
As we described before, the immediate neighbors of a node contain key attributes, and the meta-path-based neighbors can easily supply high-order connectivity information of the same node type.
Hence we choose concatenation as the COMBINE function, that is, we view the representation of the intra-type aggregation phase and the representation of the inter-type aggregation phase as the characteristics of different channels.
After the two-phase aggregation, the embedding $z_{v}$ fusing high-order intra-type information and direct inter-type information is obtained, which can be further used in different downstream tasks.
Algorithm~\ref{alg} shows the details of \mymodel{}. 

\begin{algorithm}[t]
\SetAlgoLined
\KwIn{Heterogeneous graph $\mathcal{H}$, Selected types $\mathcal{\mathcal{T}}'$, \\
 \quad\quad\quad Meta-paths $\mathcal{P} = \{\mathcal{P}_t \ | \ t \in \mathcal{T}'\}$, \\
        \quad\quad\quad Node features $\{x_v ,v \in \mathcal{V}\}$, \\
        \quad\quad\quad Number of intra-type aggregation layers $L_{intra}$,\\
        \quad\quad\quad Number of inter-type aggregation layers $L_{inter}$.\\}
\KwOut{Embeddings of target nodes}
\For{node type $t \in \mathcal{T}$}{
perform linear transformation for nodes of type $t$;
}
\tcc{Building the fused meta-path graph Using Algorithm 1} 

\tcc{Intra-type aggregation phase} 
\For{$i = 1 \to l_{intra}$}{
    \For{$t \in \mathcal{T}' $}{
        \For{$v \in \mathcal{V}_t$}{
            \For{$ u \in  N_{\mathcal{G}_t}^v$}{
                calculate learnable weight $\alpha^{t}_{uv} $ and structral semantic weight ${\tilde \delta}^{t}_{uv}$; \\
                combine $\alpha^{t}_{uv}$ and ${\tilde \delta}^{t}_{uv}$ with edge residuals; \\
            }
            calculate $h^{intra}_v$ using Equation~\ref{eq_agg};
        }
    }
}
\tcc{Inter-type aggregation phase}
\For{$i = 1 \to l_{inter}$}{
\For{$v \in \mathcal{V}$}{
 calculate $h^{inter}_{v}$ using the multi-head attention  ;\\
    }
}
\tcc{Combination of Semantic Information}
$z_{v} = h^{intra}_{v} || W_m \cdot h^{inter}_{v}$; \\
\Return $\{z_{v}, \ v \in \mathcal{V}\}$
 \caption{\mymodel{}}
 \label{alg}
\end{algorithm}

\subsection{Training}
\subsubsection{Node classification}
For the semi-supervised node classification task, we first use an MLP to adjust the dimension of node embedding to be the same as the number of classes.  
\begin{equation}
    \begin{gathered}
      z^{nc}_{v} = W_{nc} \cdot z_{v}
    \end{gathered}
    \label{eq_nc}
\end{equation}
where $W_{nc} \in \mathbb{R}^{d \times C}$. Then, for single-label classification, we use softmax to sharpen $z^{nc}_{v}$ and then employ cross-entropy as the loss function.
\begin{equation}
    \begin{gathered}
        z'_{nc} = softmax(z_{nc}) \\
      \mathcal{L} = - \sum_{v \in \mathcal{V}_L} \sum_{c = 0}^{C-1} y_{v} [c] \cdot log (z'_{nc}) [c]
    \end{gathered}
    \label{eq_nc_single}
\end{equation}
where $\mathcal{V}_L$ is the set of labeled nodes, $C$ is the number of classes, $y_v$ is the one-hot label vector of node $v$.

For multi-label classification, we apply sigmoid to $z_{nc}$ and then choose binary cross-entropy as the loss function.
\begin{equation}
    \begin{gathered}
        z'_{nc} = sigmod(z_{nc}) \\
      \mathcal{L} = - \sum_{v \in \mathcal{V}_L} \sum_{c = 0}^{C-1} l_v[c] \\
      l_v[c] = y_{v} [c] \cdot log (z'_{nc}) [c]  +  (1- y_{v} [c]) \cdot log (1 - z'_{nc}[c])
    \end{gathered}
    \label{eq_nc_multi}
\end{equation}

\subsubsection{Link prediction}
As SimpleHGN\cite{lv2021we} and R-GCN\cite{schlichtkrull2018modeling} suggests, we calculate the probability that nodes $u$ and $v$ are connected by edge $e$ with type $r$.
\begin{equation}
    \begin{gathered}
      prob(e) = prob(u,v) = sigmod(z_{v}^T \cdot W_{r} \cdot z_{u}) \\
        \mathcal{L} = - \sum_{e \in \mathcal{E}_L} y_{e} \cdot log (prob(e)) + (1 - y_{e}) \cdot log (1-prob(e))
    \end{gathered}
    \label{eq_lp}
\end{equation}
where $\mathcal{E}_L$ is the edge set for model training and $W_{r} \in \mathbb{R}^{ d \times d}$.
\subsection{Complexity Analysis}
\label{ss:complexity}
In this section, we theoretically analyze the complexity of \mymodel{} during the training stage and compare it with the typical meta-path-based model HAN and meta-path-free model SimpleHGN.
Suppose the number of nodes in the heterogeneous graph is $N$, the number of edges is $E$, the dimension of the node raw feature is $D$, the node embedding dimension after linear transformation is $ D'$.
For the fused meta-path graph, the number of edges is $E'$ and the number of nodes is $N’$.
\subsubsection{Time complexity}
For \mymodel{}, in the linear transformation phase and  semantic information combination phase, the complexities are all $O(NDD')$. In the intra-type aggregation phase, attention is calculated pairwise between nodes in the fused meta-path graph, its complexity is  
$O(E’D’+N’D’^2)$. The inter-type aggregation phase is carried out in the original graph, the time complexity is $O(ED’+ND’^2)$, which is consistent with GAT. 

For SimpleHGN, the complexity of its linear transformation phase is $O(NDD')$.
Since it introduces edge features, let the dimension of edge features be $F$, the number of edge types be $T$, the complexity of calculating attention and aggregating neighbors is $O(E(D’+F) +ND’^2+TF^2)$. 

Overall, the complexity of \mymodel{} is $O(NDD’) + O(E’D’+ED’+ND’^2)$. The complexity of SimpleHGN is $O(NDD’) + O(TF^2+EF+ED’+ND’^2)$. 
It can be seen that the complexity of the two models is similar. The difference is mainly in the intra-type aggregation phase of \mymodel{}, and the edge feature usage of SimpleHGN. 

Meanwhile, the complexity of HAN is $O(K\epsilon D’^2 + KD’^2 +ND’^2)$, where $K$ is the number of meta-paths, $\epsilon$ is the number of edges in each meta-path-based graph. 
Thanks to the fused meta-path graph, \mymodel{} can reduce redundant edges and avoid double-layer attention.
\subsubsection{Space complexity}
For \mymodel{}, 
in the intra-type aggregation and inter-type aggregation phases, attentions over edges are calculated. 
The dimensions of the parameter matrix $W$ and the parameter vector $a^T$ are $O(2\times D' \times D')$ and $O(D')$, respectively. 
The dimensions of attention weights $\alpha$ are $O(E')$ in the intra-type aggregation phase, and $O(E)$ in the inter-type aggregation phase. 
Thus, the overall space complexity of \mymodel{} is $O(D'^2 + E' + E)$.

For SimpleHGN, since it introduces edge features, the space complexity is $O(D'^2 + E + TF^2)$. 
For HAN, its space complexity is $O(K(D'^2 + \epsilon))$. Since the fused meta-path graphs employ a union operation to handle all meta-path graphs (i.e., $K \times \epsilon$ is much larger than $E'$), the memory cost of \mymodel{} is less than SimpleHGN and HAN.
%
%
%

Experimental results in Section~\ref{ss:efficiency} reveal that \mymodel{} can achieve better performance with higher efficiency than existing HGNNs from the following perspectives, i.e., parameter size, FLOPs, memory overhead, and runtime per training epoch.

\section{Experiments}

In this section, we conduct extensive experiments to answer the following questions:
\begin{itemize}
 \item {\bfseries RQ1}: How is the effectiveness of the proposed \mymodel{} compared with existing heterogeneous GNN models?
 \item {\bfseries RQ2}: What is the impact of each major component of \mymodel{}?
 \item {\bfseries RQ3}: How about the  efficiency of \mymodel{}?
  \item {\bfseries RQ4}: How to evaluate the quality of node representations learned by \mymodel{} in a visual way?
 \item {\bfseries RQ5}: How robust is \mymodel{} to hyperparameter?
 \item {\bfseries RQ6}: Are meta-paths or variants still useful in heterogeneous GNNs~\cite{lv2021we}? How to select suitable meta-paths?
\end{itemize}

\subsection{Experimental Setup}
\subsubsection{Experimental Setting}
All experiments are conducted under the recently proposed Heterogeneous Graph Benchmark (HGB)~\cite{lv2021we}. 
HGB provides unified data split and data preprocessing to ensure the fairness of comparison.
In the node classification task, node labels are split according to 24\% for training, 6\% for validation, and 70\% for test in each dataset. 
In the link prediction task, the test set uses 2-hop neighbors as negative. 
%
%
To prevent data leakage, the evaluation metrics are obtained by submitting predictions to the HGB website\footnote{https://www.biendata.xyz/competition/hgb-1/}.
All experiments are run on a single GPU (NVIDIA
Tesla V100) with 32 GB memory.

\begin{table}[t]
  \caption{Statistics of the datasets.}
  \label{Statistics of the datasets}
  \centering
  \resizebox{\columnwidth}{!}{
  \begin{tabular}{ccclcc}
    \toprule
    Node Classification     & \#Nodes     & \#Node Types  & \#Edges &Target & \#Classes \\
    \midrule
    DBLP & 26128  & 4  & 239566  & author & 4   \\
    IMDB & 21420 & 4 & 86642  & movie & 5   \\
    Freebase & 180098 & 8 & 1057688  & book & 7   \\
    \midrule
    Link Prediction &    &  &    & \multicolumn{2}{c}{Target}   \\
    \midrule
    LastFM & 20612  & 3& 141521  & \multicolumn{2}{c}{user-artist}   \\
    PubMed & 63109  & 4  &244989  &\multicolumn{2}{c}{disease-disease}   \\
    \bottomrule
  \end{tabular}}
  \vspace{-2ex}
\end{table}

\begin{table*}[ht]
  \caption{Performance comparison on node classification (the higher, the better). The bold and the underline indicate the best and the second best. ‘OOM’ means out of memory.* indicates the result is
statistically significant (t-test with p-value  \textless 0.01)}
  \centering
  \scalebox{0.95}{
  \label{Node Classification}
  \begin{tabular}{l|cc|cc|cc}
    \toprule
    Dataset   & \multicolumn{2}{c}{DBLP}    & \multicolumn{2}{c}{IMDB} & \multicolumn{2}{c}{Freebase} \\
    \midrule
    Model $\backslash$ Metrics &  Macro-F1 & Micro-F1 & Macro-F1 & Micro-F1& Macro-F1 & Micro-F1 \\
    \midrule
    HAN & 93.17$\pm$0.19	& 93.64$\pm$0.17	 &	59.70$\pm$0.90 &	65.61$\pm$0.54 & 21.31$\pm$1.68	& 54.77$\pm$1.40\\
    BPHGNN & 93.89$\pm$0.43	& 94.50$\pm$0.47	 &	64.01$\pm$0.56 &	67.77$\pm$0.70 & OOM	&OOM\\
    GTN & 65.21$\pm$1.16	& 66.23$\pm$1.23	&	59.26$\pm$0.84	& 64.07$\pm$0.65 & OOM	& OOM\\
    MAGNN & 93.16$\pm$0.38 &	93.65$\pm$0.34 &		56.92$\pm$1.76 &	65.11$\pm$0.59& OOM	& OOM \\
    HetSANN& 84.08$\pm$1.01&	84.96$\pm$0.88&	49.25$\pm$0.57&	57.47$\pm$1.12& OOM	& OOM  \\
    HGNN-AC &92.97$\pm$0.72&	93.43$\pm$0.96&56.63$\pm$0.81&	63.85$\pm$0.85& OOM	& OOM \\
    R-HGNN & 93.31$\pm$0.42 & 93.80$\pm$0.39&61.39$\pm$0.51&	66.03$\pm$0.65& 46.29$\pm$1.21&	59.43$\pm$0.95 \\
    CKD & 92.52$\pm$0.23 & 92.80$\pm$0.22 &60.30$\pm$0.77 & 65.98$\pm$0.87& 45.88$\pm$1.53 &59.18$\pm$0.63 \\
    \midrule
    RGCN & 91.52$\pm$0.50	& 92.07$\pm$0.50 & 58.85$\pm$0.26 & 62.05$\pm$0.15& 46.78$\pm$0.77	& 58.33$\pm$1.57 \\
    HGT & 92.77$\pm$0.35	& 93.44$\pm$0.31 & 63.02$\pm$0.80 & 67.01$\pm$0.36& 29.28$\pm$2.52	& 60.51$\pm$1.16 \\
    HetGNN &92.77$\pm$0.24&	93.23$\pm$0.23&	47.87$\pm$0.33&	50.83$\pm$0.26& OOM	& OOM \\
    GCN& 90.54$\pm$0.27&	91.18$\pm$0.25	&	59.95$\pm$0.72&	65.35$\pm$0.35& 27.84$\pm$3.13	& 60.23$\pm$0.92 \\
    GAT &92.96$\pm$0.35&	93.46$\pm$0.35&		56.95$\pm$1.55&	64.24$\pm$0.55& 40.74$\pm$2.58	& 65.26$\pm$0.80 \\
    AMHGNN &93.71$\pm$0.79&	 94.08$\pm$0.76&		63.38$\pm$0.66&	67.29$\pm$0.64& 47.12$\pm$1.65	& 65.55$\pm$1.89 \\
    HINormer &\underline{94.57$\pm$0.21}&	\underline{94.94$\pm$0.23}&		\underline{64.65$\pm$0.53}&	\underline{67.83$\pm$0.34}& OOM	& OOM \\
    SimpleHGN& 93.81$\pm$0.54&	94.26$\pm$0.49&	63.53$\pm$1.66&	67.42$\pm$0.42& \underline{47.57$\pm$1.23}	& \underline{65.89$\pm$0.50} \\
    \midrule
    \textbf{\mymodel{}} &\textbf{95.06$\pm$0.31\*}&	\textbf{95.40$\pm$0.30*}&	\textbf{65.57$\pm$0.49*}&	\textbf{68.62$\pm$0.23*}& \textbf{49.57$\pm$1.50*}	& \textbf{66.31$\pm$0.27*} \\ 
    \bottomrule
    \end{tabular}}
    \vspace{-1ex}
\end{table*}

\subsubsection{Datasets}
We use five real-world datasets for node classification, node clustering, and link prediction.
The statistics of datasets are summarized in Table~\ref{Statistics of the datasets}.

\begin{itemize}
    \item \textbf{DBLP} is a computer science bibliography website containing author (A), paper (P), term (T), and conference (C). 
    \item \textbf{IMDB} is a movie website, which contains movie (M), director (D), actor (A), and keyword (K).
    \item \textbf{Freebase} is a huge knowledge graph with book (B), film (F), music (M), organization (O), business (U), etc.
    \item \textbf{LastFM} is an online music website containing user (U), artist (A), and tag (T). The target edge type is user-artist. 
     \item \textbf{PubMed} is a biomedical literature library, which has gene (G), disease (D), chemical (C), and specie (S). The target is to predict the connection between diseases.
\end{itemize}

\subsubsection{Implementation Details}
The parameters are randomly initialized. 
We use Adam~\cite{kingma2014adam} to optimize parameters. 
$\beta$ in intra-type aggregation is set to 0.3. 
We set the number of intra-type aggregation layers to 2 for all datasets, the number of inter-type aggregation layers to 5 for the IMDB dataset and 2 for other datasets.
All GNN models are implemented with PyTorch.
The selected meta-paths are listed in Table~\ref{Metapaths selected} (Section~\ref{ss:meta-path}).

\subsection{Performance Comparison (RQ1)}
\subsubsection{Node Classification}
We select baselines depending on whether or not the meta-path is used.
\begin{itemize}
    \item \textbf{Meta-path-based:} HAN~\cite{wang2019heterogeneous}, GTN~\cite{yun2019graph},
    MAGNN~\cite{fu2020magnn},  HetSANN~\cite{hong2020attention},
    HGNN-AC~\cite{jin2021heterogeneous}, 
    R-HGNN~\cite{yu2022heterogeneous}, CKD~\cite{wang2022collaborative} and
    BPHGNN~\cite{fu2023multiplex}.
    \item \textbf{Meta-path-free:} R-GCN\cite{schlichtkrull2018modeling}, HGT~\cite{hu2020heterogeneous}, HetGNN~\cite{zhang2019heterogeneous}, GCN~\cite{kipf2016semi},
    GAT~\cite{velivckovic2017graph},
    SimpleHGN~\cite{lv2021we},
    AMHGNN~\cite{li2024node},
    and HINormer~\cite{mao2023hinormer}.
\end{itemize}

We run these baselines using the official codes. The hyperparameter settings is consistent with HGB.
%
We adopt the Macro F1 and Micro F1 metrics for node classification.
All models are run five times and the mean and standard deviation are reported.
%
The results are shown in Table~\ref{Node Classification}. 

%
%
Neither the meta-path-based models or the meta-path-free models can always achieve better performance on all datasets. 
For meta-path-based models, HAN, MAGNN, and R-HGNN performs better on the DBLP dataset. 
R-HGNN and CKD performs better on the IMDB dataset.
Nevertheless, in contrast with the meta-path-based models, the 
meta-path-free models such as the commonly used GAT and GCN can also achieve competitive or even better performance. 
%
%
%
%
Also, SimpleHGN serves as a strong baseline, indicating that it is necessary to fuse different types of information in the node classification task.
The results show that intra-type aggregation and inter-type aggregation have their own advantages and disadvantages.
For relatively large datasets such as Freebase, existing models run out of memory due to either focusing too much on meta-path (e.g., MAGNN, GTN, HetSANN) or too high model complexity (e.g., HetGNN).

In comparison, \mymodel{} can not only run on large datasets but also consistently achieves the best performance on all three datasets, which demonstrates that intra-type and inter-type aggregations are complementary and making good use of type information in heterogeneous graphs is crucial.

\subsubsection{Node Clustering}

\begin{table}
  \caption{Performance comparison on Node Clustering (the higher, the better). * indicates the result is
statistically significant (t-test with p-value \textless 0.01).}
  \resizebox{\columnwidth}{!}{
  \label{Node Clustering}
  \centering
  \begin{tabular}{l|cc|cc}
    \toprule
    Dataset   & \multicolumn{2}{c}{DBLP}  & \multicolumn{2}{c}{IMDB}  \\
    \midrule
    Model $\backslash$ Metrics &NMI & ARI & NMI & ARI  \\
    \midrule
    HAN & 72.98$\pm$2.64	& 78.50$\pm$2.80	& 15.34$\pm$0.36	& 11.06$\pm$0.29  \\ 
    MAGNN & 79.88$\pm$0.88	& 85.38$\pm$0.75	& 15.76$\pm$0.53	& 12.82$\pm$0.63  \\
    RHGNN & 57.21$\pm$0.09	& 77.68$\pm$0.17	& 15.29$\pm$0.38	& 12.26$\pm$0.44  \\
    \midrule
    RGCN & 72.32$\pm$1.77	& 78.12$\pm$2.20	& 7.63$\pm$0.45	& 5.54$\pm$0.4  \\
    GCN & 71.88$\pm$0.71	& 78.67$\pm$0.77	& 15.71$\pm$0.32	& 12.31$\pm$0.57  \\
    GAT & 77.15$\pm$1.69	& 82.00$\pm$2.28	& 15.42$\pm$0.96	& 12.13$\pm$1.30  \\
    SimpleHGN 	 &	 80.55$\pm$1.15 & 	86.11$\pm$1.10 &	15.95$\pm$1.38 &	12.79$\pm$0.95 \\    
    
    \textbf{\mymodel{}} 	 &	 \textbf{83.51$\pm$0.51*} & \textbf{88.51$\pm$0.53*} &	\textbf{16.44$\pm$1.05*} &	\textbf{13.05$\pm$0.86*} \\
    \bottomrule
  \end{tabular}}
  \vspace{-2ex}
\end{table}
To verify the quality of node embeddings generated by different models, we conduct node clustering on the IMDB and DBLP datasets.
The labeled nodes (i.e., movies in IMDB and authors in DBLP) are clustered  with the $k$-means algorithm. 
The number of clusters in $k$-means is set to the number of classes for each dataset, i.e., 3 for IMDB and 4 for DBLP.
We employ the normalized mutual information (NMI) and the adjusted rand index (ARI) as evaluation metrics.
From Table~\ref{Node Clustering}, we see that \mymodel{} regularly outperforms all other baselines in the node clustering task. 
Note that all models perform significantly worse on IMDB than on DBLP. 
This is due to the dirty labels of movies in IMDB, i.e., every movie node in the original IMDB dataset has multiple genres, and we only choose the very first one as its class label~\cite{fu2020magnn}.
As shown in Table~\ref{Node Clustering}, the traditional heterogeneous models do not have many advantages over the traditional homogeneous models in node clustering.
And the node embedding generated by \mymodel{} has higher quality, leading to better clustering effect.
\subsubsection{Link Prediction}

\begin{table}
  \caption{Performance comparison on link prediction. ‘OOM’ means out of memory. * indicates the result is
statistically significant (t-test with p-value \textless 0.01).}
  \resizebox{\columnwidth}{!}{
  \label{link prediction}
  \centering
  \begin{tabular}{l|cc|cc}
    \toprule
    Dataset   & \multicolumn{2}{c}{LastFM}  & \multicolumn{2}{c}{PubMed}  \\
    \midrule
    Model $\backslash$ Metrics &ROC-AUC & MRR & ROC-AUC & MRR  \\
    \midrule
    MAGNN & 56.81$\pm$0.05	& 72.93$\pm$0.59	& OOM	& OOM  \\
    \midrule
    RGCN & 57.21$\pm$0.09	& 77.68$\pm$0.17	& 78.29$\pm$0.18	& 90.26$\pm$0.24  \\
    GATNE & 66.87$\pm$0.16	& 85.93$\pm$0.63	& 63.39$\pm$0.65	& 80.05$\pm$0.22  \\ 
    HetGNN & 62.09$\pm$0.01	& 85.56$\pm$0.14	& 73.63$\pm$0.01	& 84.00$\pm$0.04  \\
    HGT & 54.99$\pm$0.28	& 74.96$\pm$1.46	& 80.12$\pm$0.93	& 90.85$\pm$0.33  \\
    GCN & 59.17$\pm$0.31	& 79.38$\pm$0.65	& 80.48$\pm$0.81	& 90.99$\pm$0.56  \\
    GAT & 58.56$\pm$0.66	& 77.04$\pm$2.11	& 78.05$\pm$1.77	& 90.02$\pm$0.53  \\
    SimpleHGN 	 &	 \underline{67.16$\pm$0.37} & 	\underline{86.73$\pm$0.27} &	\underline{83.39$\pm$0.39} &	\underline{92.07 $\pm$0.26} \\  
    AutoAC 	 &	 66.64$\pm$0.26 & 	86.16$\pm$0.51 &	82.95$\pm$0.38 &	91.68$\pm$0.35 \\    
    \midrule
    \textbf{\mymodel{}} 	 &	 \textbf{67.33$\pm$0.10*} & \textbf{86.84$\pm$0.14*} &	\textbf{84.00$\pm$0.63*} &	\textbf{94.63$\pm$0.29*} \\
    
    \bottomrule
  \end{tabular}}
  \vspace{-3ex}
\end{table}

\begin{table*}[ht]
  \caption{Ablation study on \mymodel{}. DBLP and IMDB is for node classification and PubMed is for link prediction.}
   \centering
  \scalebox{0.95}{
  \label{ablation Study}
 
  \begin{tabular}{l|cc|cc|cc}
    \toprule
    Dataset   & \multicolumn{2}{c}{DBLP}  & \multicolumn{2}{c}{IMDB}  & \multicolumn{2}{c}{PubMed}  \\
    \midrule
    Model $\backslash$ Metrics & Macro-F1 & Micro-F1 & Macro-F1 & Micro-F1 & ROC-AUC & MRR  \\
    \midrule
    HAGNN-wo-inter   &	93.85$\pm$0.53& 93.27$\pm$0.64 &	57.03$\pm$2.60 &	63.46$\pm$1.17 &	 81.89$\pm$0.07 & 93.79$\pm$0.18 \\
    HAGNN-wo-intra & 93.11$\pm$0.69	& 93.61$\pm$0.63 & 57.34$\pm$0.87 &	62.62$\pm$0.49 &  72.18$\pm$0.17 &	89.38$\pm$0.30 \\
    HAGNN-wo-sw& 94.52$\pm$0.01	& 94.62$\pm$0.08&	65.08$\pm$0.44&	67.90$\pm$0.19&	83.49$\pm$0.22&	94.31$\pm$1.12  \\
    HAGNN-wo-fused & 94.22$\pm$0.28	& 94.59$\pm$0.28& 60.91$\pm$2.15 &	63.21$\pm$2.13&81.99$\pm$2.47&	93.06$\pm$0.55 \\
    HAGNN-wo-combine & 94.77$\pm$0.22	& 95.15$\pm$0.19& 65.31$\pm$0.23 &	68.11$\pm$0.20&82.70$\pm$0.69&	92.37$\pm$0.34 \\
    HAGNN-combine-add & 95.00$\pm$0.19& 95.25$\pm$0.27	& 65.32$\pm$0.33 &	68.21$\pm$0.30&83.10$\pm$0.51&	93.17$\pm$0.44 \\
    HAGNN-reverse & 94.04$\pm$0.77& 94.37$\pm$0.68	& 58.88$\pm$1.33 &	63.32$\pm$1.06&81.97$\pm$0.21&	93.57$\pm$0.24\\
    \textbf{HAGNN} &\textbf{95.06$\pm$0.30}&	\textbf{95.40$\pm$0.30}&	\textbf{65.57$\pm$0.49}&	\textbf{68.62$\pm$0.23}&	\textbf{84.00$\pm$0.30}&	\textbf{94.63$\pm$0.29} \\ 
    
    \bottomrule
    \end{tabular}}
    \vspace{-2ex}
\end{table*}

We select widely used link prediction models as baselines, including RGCN, GATNE~\cite{cen2019representation}, HetGNN, MAGNN, HGT, GCN, GAT, SimpleHGN, and AutoAC~\cite{zhu2023autoac}. 
The link prediction task is performed on the LastFM and Pubmed datasets.
We adopt the MRR and ROC-AUC metrics and the mean and standard deviation of five runs are reported.
From Table~\ref{link prediction}, we see that GCN, GAT, and other direct neighbor aggregation models perform better than MAGNN and HGT on the LastFM dataset.
This is mainly due to the weak heterogeneity of LastFM, which contains only three node types and three edge types. 
The heterogeneous graph models on the PubMed dataset can achieve comparable results. 
In comparison, \mymodel{} outperforms existing models on both datasets, especially on PubMed, where the MRR metric improves by 2\% compared to SimpleHGN.  
%

Overall, \mymodel{} achieves better performance in different tasks and different datasets.
Moreover, \mymodel{} can easily handle larger datasets.

\subsubsection{Comparison on datasets of other real-world domains}
Recent works~\cite{liu2023datasets,han2022openhgnn} propose two datasets from new real-world domains.
\begin{itemize}
    \item \textbf{Risk Commodity Detection Dataset (RCDD)}: RCDD is based on a real risk detection scenario from Alibaba’s e-commerce platform with 157,814,864 edges and 13,806,619 nodes.
    The target node type is \emph{item}.
    For confidentiality and security, other six node types are represented by single letters (i.e., \emph{a}, \emph{b}, \emph{c}, \emph{d}, \emph{e}, \emph{f}). 
   Due to the immense scale of RCDD, even the simple HGNN models in OpenHGNN~\footnote{\url{https://github.com/BUPT-GAMMA/OpenHGNN}} fail to run due to OOM. 
   For fair comparison, we perform sampling  on the original dataset.
   Specifically, for node types \emph{item}, \emph{a}, \emph{e}, and \emph{f}, we randomly sample 100,000 nodes for each type, while we retain all nodes for type \emph{b} and type \emph{d}.
   The sampled sub-graph contains 447,289 nodes in total.
   
    \item \textbf{Takeout Recommendation Dataset (TRD)}, which is collected from 11 commercial districts in Beijing from March 1st to March 28th, 2021. It has three types, i.e., \emph{spu}, \emph{poi}, and \emph{user}, where \emph{poi} is the takeout restaurant, and \emph{spu} is food, and this graph is huge with 18,931,400 edges and 408,849 nodes.
    The graph task is link prediction, predicting whether there is an edge between food and the user. 
    The evaluation metric is AUC-ROC.

\end{itemize}

%
Table~\ref{Comparison on Large-scale Dataset} shows the performance comparison on RCDD and TRD.
\mymodel{} still achieves the best performance on the two datasets from different domains, which further validates the effectiveness of \mymodel{}.
\begin{table}
  \caption{Performance comparison on RCDD and TRD ("-" means the model can not run on the link prediction task)}
  \resizebox{0.9\columnwidth}{!}{
  \label{Comparison on Large-scale Dataset}
  \centering
  \begin{tabular}{l|cc|c}
    \toprule
    Dataset   & \multicolumn{2}{c}{RCDD}  & TRD  \\
    \midrule
    Model $\backslash$ Metrics &Macro-F1 & Micro-F1 & ROC-AUC  \\
    \midrule
    HAN & 74.59$\pm$1.03	& 79.06$\pm$1.12	& 89.33$\pm$0.77  \\ 
    HetSANN & 70.30$\pm$0.70	& 75.06$\pm$0.86	& -  \\
    \midrule
    RGCN & \underline{81.36$\pm$0.55}	& 85.04$\pm$0.56	& 92.63$\pm$0.41	  \\
    GCN & 72.88$\pm$1.09	& 78.76$\pm$0.98	& 90.55$\pm$0.96	  \\
    GAT & 77.15$\pm$1.69	& 82.00$\pm$2.28	& 90.73$\pm$0.88	 \\
    RGAT & 79.85$\pm$0.91	& 86.00$\pm$0.93	& 91.05$\pm$0.65	 \\
    SimpleHGN 	 &	 80.55$\pm$1.15 & 	\underline{86.11$\pm$1.10} &	\underline{92.60$\pm$0.82}  \\    
    \textbf{\mymodel{}} 	 &	 \textbf{82.34$\pm$0.74} & \textbf{87.42$\pm$0.80} &	\textbf{93.11$\pm$0.63}  \\
    \bottomrule
  \end{tabular}}
  \vspace{-2ex}
\end{table}
\subsection{Ablation Study (RQ2)}
We design the following variants of \mymodel{}.
\begin{itemize}
    \item \textbf{HAGNN-wo-inter} removes the inter-type aggregation phase and directly performs downstream tasks on the embeddings obtained after the intra-type aggregation phase.
    \item \textbf{HAGNN-wo-intra} removes the intra-type aggregation phase and aggregates the direct neighbors of all nodes.
    \item \textbf{HAGNN-wo-sw} follows the two-phase framework, but in the intra-type aggregation phase, the semantic structure information is not utilized.
    \item \textbf{HAGNN-wo-fused} conducts intra-type aggregation on meta-path-based graphs rather than the fused meta-path graph.
    \item \textbf{HAGNN-wo-combine} removes the combination of the embedding obtained in intra-type aggregation and inter-type aggregation.
    \item \textbf{HAGNN-combine-add} uses the \emph{add} operation to combine the the embedding obtained in intra-type aggregation and inter-type aggregation.
    \item \textbf{HAGNN-reverse} change the order of HAGNN, which conducts inter-type aggregation first, then the intra-type aggregation.

\end{itemize}

Table~\ref{ablation Study} shows the performance comparison between \mymodel{} and its six variants on the node classification (i.e, DBLP and IMDB) and link prediction (i.e., PubMed) tasks. 
\mymodel{}-wo-intra performs better than \mymodel{}-wo-inter on the DBLP dataset. 
While HAGNN-wo-inter is much better than HAGNN-wo-intra on the PubMed dataset.
No matter which variant, its performance is far inferior to \mymodel{}.
That is, both the higher-order intra-type information and direct inter-type information to the target nodes are important, but their importance varies in different datasets. 
%
%
The proposed two-phase aggregation method in \mymodel{} can leverage the information within and between types simultaneously.

Moreover, the performance of \mymodel{}-wo-sw is inferior to that of \mymodel{}, which indicates that the proposed structural semantic weight helps to aggregate intra-type information. 
Also, \mymodel{}-wo-fused performs worse than \mymodel{}, especially on the  IMDB and PubMed datasets. 
The main reason is that the information redundancy between multiple meta-path-based graphs plays an negative effect on the performance.
\mymodel{}-wo-combine and \mymodel{}-combine-add perform worse than \mymodel{}, showing that the concatenation combination can work better.
Additionally, \mymodel{}-reverse can not work as good as \mymodel{}, indicating that it's necessary to do intra-type aggregation first. 

\subsection{Efficiency Comparison (RQ3)}
\label{ss:efficiency}

We further evaluate and compare the efficiency of \mymodel{} and SimpleHGN from the following perspectives, i.e., parameter size, FLOPs, memory overhead, and runtime per training epoch. 
Let HAN-\mymodel{} denote the heterogeneous GNN that uses HAN's meta-path-based graphs instead of the proposed fused meta-path graph in the intra-type aggregation phase. 
From Table~\ref{time complexity}, we see that \mymodel{} achieves the best efficiency among the three models, which is consistent with the complexity analysis in Section~\ref{ss:complexity}.

Moreover, due to less parameters and FLOPs during the training process, \mymodel{} is also faster than SimpleHGN and HAN-\mymodel{}.
%
Although \mymodel{} is composed of two phases (i.e., intra-type and inter-type aggregation phases), \mymodel{} can achieve better performance with higher efficiency than existing models with one aggregation phase due to the fused meta-path graph and structural semantic aware aggregation mechanism. 
%
\begin{table}[t]
  \caption{Efficiency comparison between SimpleHGN and \mymodel{}.}
  \resizebox{\columnwidth}{!}{
  \label{time complexity}
  \centering
  \begin{tabular}{c|cc|cc|cc|cc}
    \toprule
   \multirow{2}{*}{\makecell{Model}} & \multicolumn{2}{c}{\makecell{\#Params \\(Million)}} & \multicolumn{2}{c}{\makecell{\#FLOPs \\ (Billion)}} & \multicolumn{2}{c}
   {\makecell{\#Memory \\ (MB)}} & \multicolumn{2}{c}{\makecell{Runtime per Training Epoch \\ (GPU Seconds)}} \\
    \cline{2-9}
    & DBLP & IMDB & DBLP & IMDB & DBLP & IMDB&DBLP & IMDB \\
    \midrule

    SimpleHGN & 2.303  &3.464 & 51.172 & 70.869 &4472&3088& 0.201 & 0.170 \\
    HAN-\mymodel{} & 3.118  &4.419 & 37.392 & 59.636 &5306&4128& 0.342 & 0.270 \\
    \mymodel{} & \textbf{2.012}  & \textbf{3.313} & \textbf{31.888} & \textbf{53.007}&\textbf{4209}&\textbf{2822}& \textbf{0.161} & \textbf{0.146} \\
    \bottomrule
    \end{tabular}}
\end{table}
\subsection{Visualization  (RQ4)}

\begin{figure}
  \centering
      \subfigure[MAGNN]{
    \includegraphics[width=0.25\columnwidth]{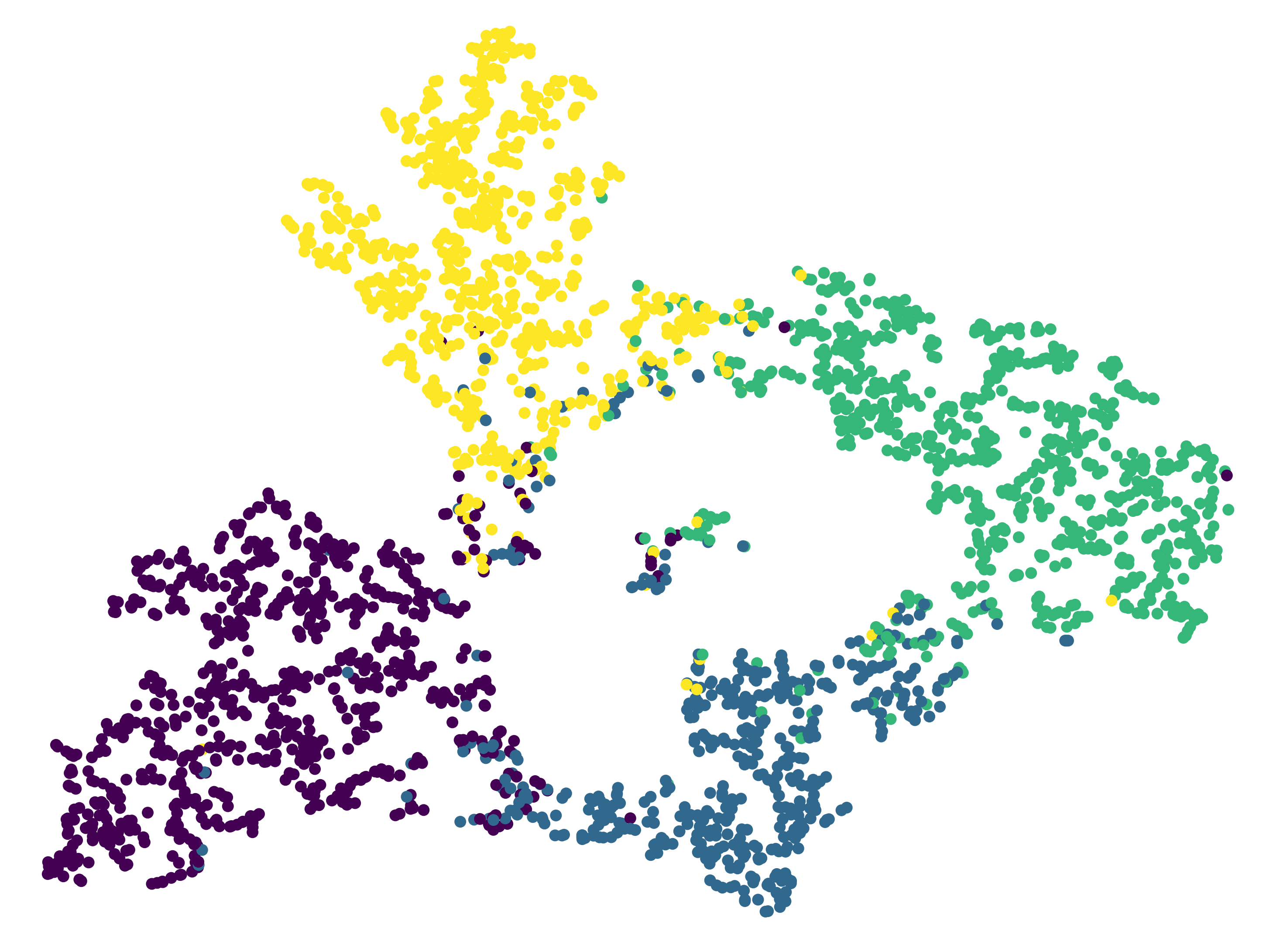}
    }
    \hspace{0.0005\columnwidth}
    \subfigure[GAT]{
    \includegraphics[width=0.25\columnwidth]{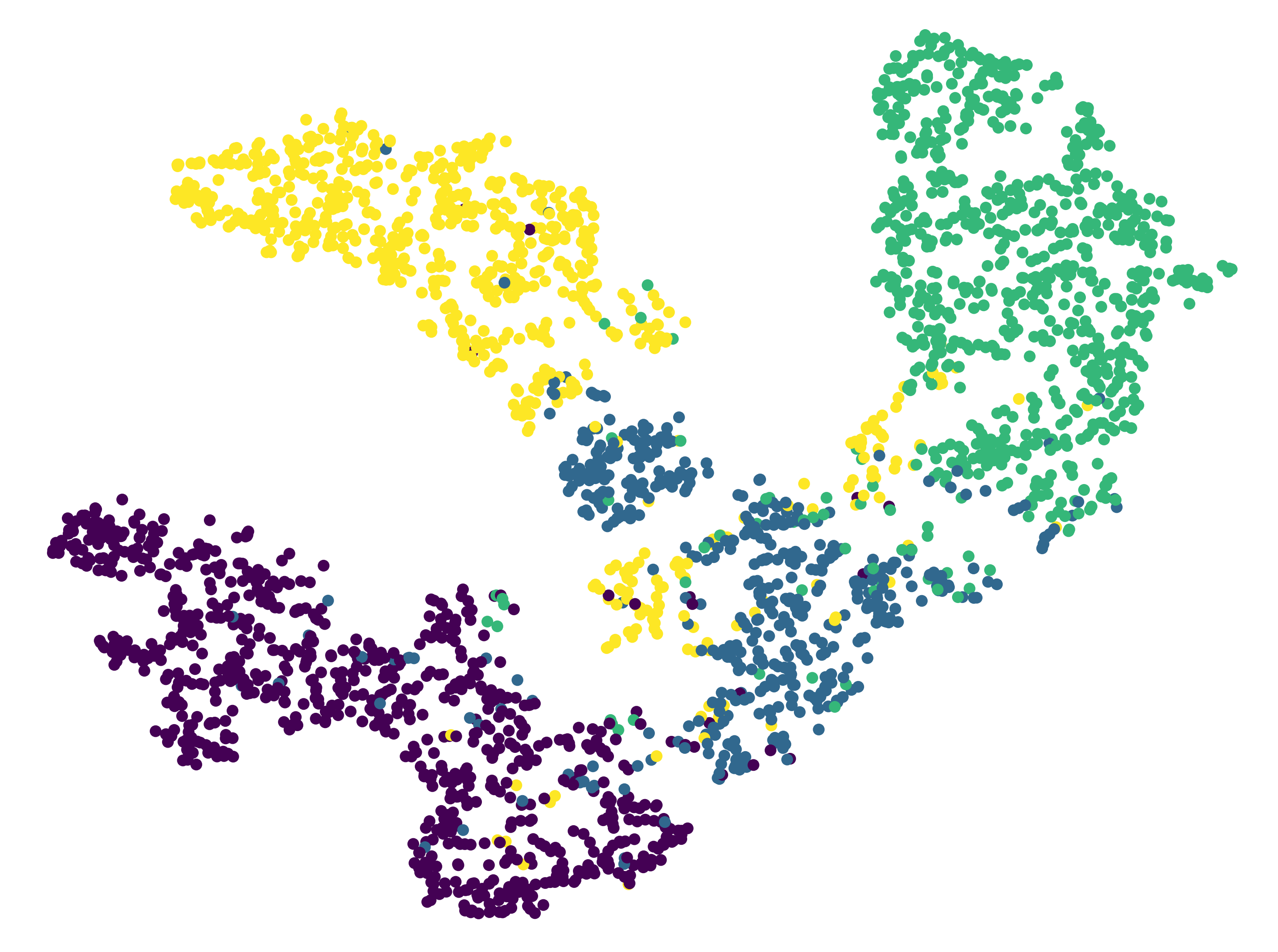}
    }
    \hspace{0.0005\columnwidth}
    \subfigure[HAGNN]{
    \includegraphics[width=0.25\columnwidth]{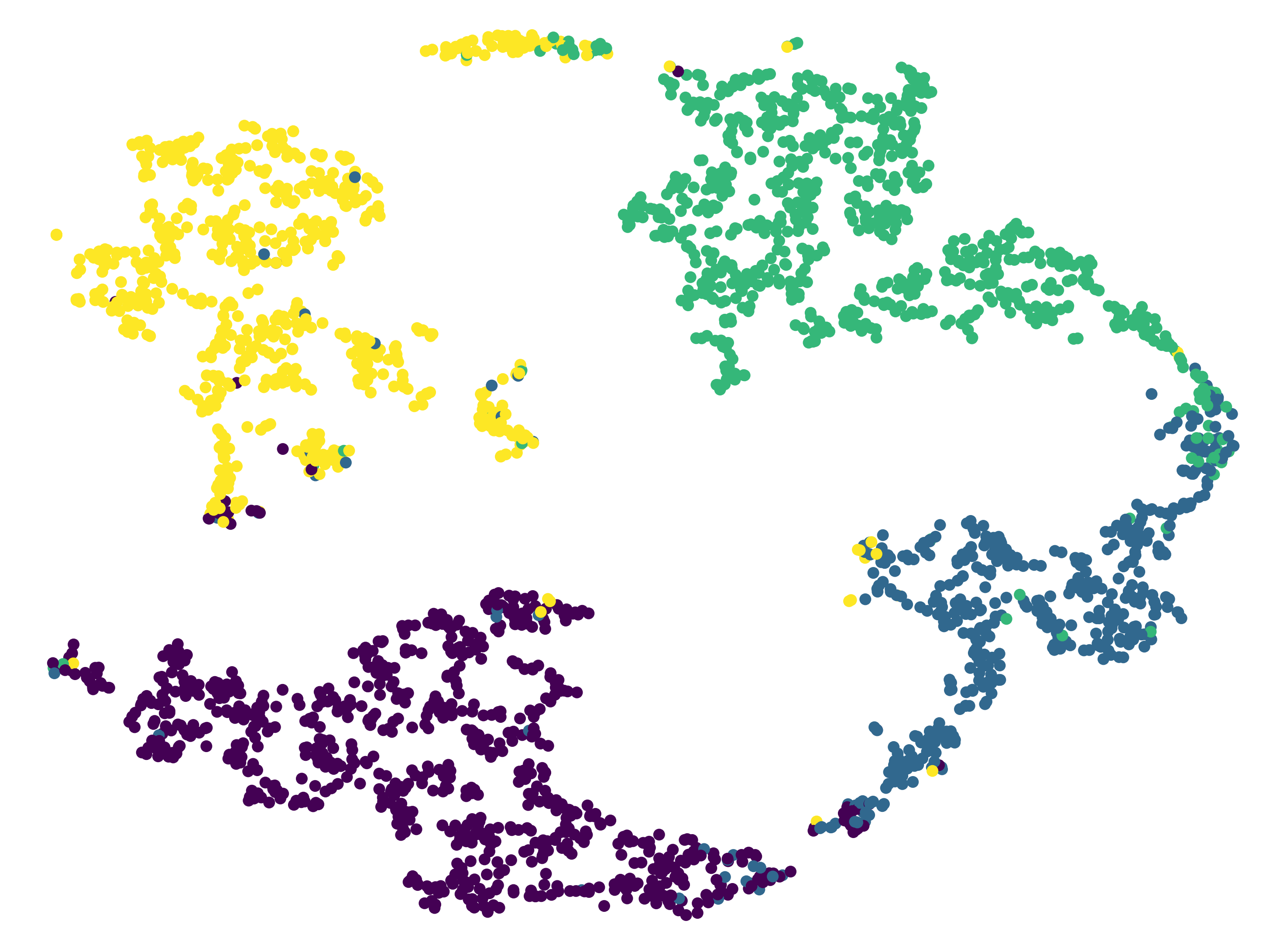}
    }
  \caption{Node visualization on the DBLP dataset.}
  \label{visualization results on DBLP dataset}
  \vspace{-3ex}
\end{figure}
\subsubsection{Quality of node representation}
%
To demonstrate the quality of node representations, we project the low-dimensional node embeddings into the two-dimensional space using t-SNE~\cite{van2008visualizing}, and the visualization results are shown in Figure~\ref{visualization results on DBLP dataset}. 
Different colors represent different classes. 
We can see that \mymodel{} can generate more clear classification boundaries than commonly used models (i.e., MAGNN and GAT).
Moreover, the point distribution in the same class is more close, indicating that the node embedding generated by \mymodel{} has higher quality.
%
The clustering results of other baselines can be seen in Table \ref{Node Clustering}.
Overall, visualization results further demonstrate the effectiveness of \mymodel{}.

\subsubsection{Difference Between Intra-type Aggregation and Inter-type Aggregation}
\begin{figure}
	\centering
	\includegraphics[width=0.95\columnwidth]{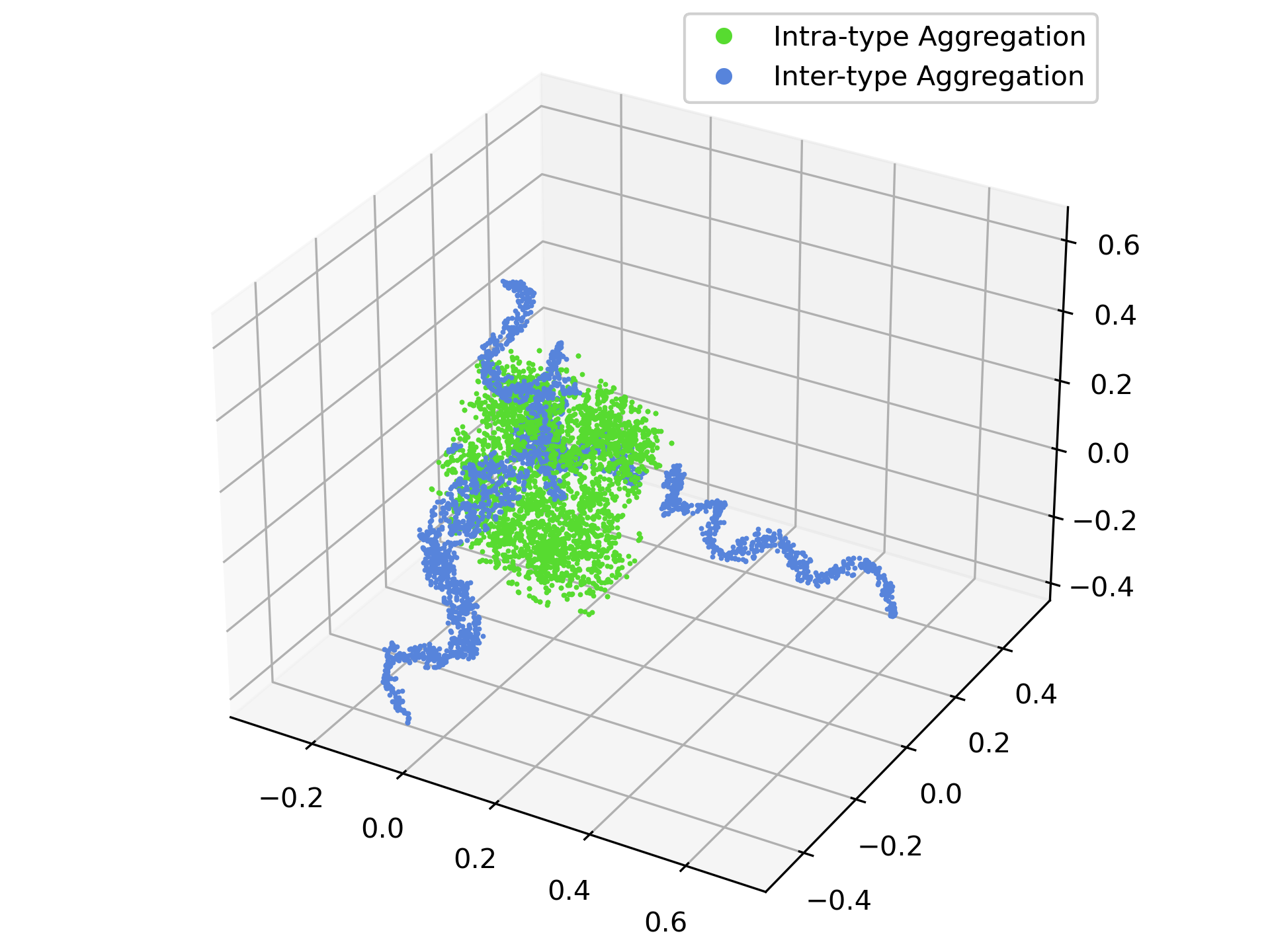}
	\caption
	{
		Node embedding visualization of intra-type and inter-type aggregation phases on the DBLP dataset.
	}
	\label{fig:compare2pahse}
\end{figure}
Due to the different neighborhoods selected in the aggregation phase, intra-type aggregation and inter-type aggregation actually extract the semantic information encoded in heterogeneous graphs from different perspective.
To further verify the difference between the information obtained in the two phases, we project node embeddings of the target type generated by each phase into a three-dimensional space using t-SNE. 
The result is illustrated in Figure~\ref{fig:compare2pahse}.
We see that the node embedding of intra-type aggregation is denser while the node embedding of inter-type aggregation is looser.
More importantly, the two types of embeddings are distributed on different planes, which indicates that they extract the semantic information from different perspectives. 

\begin{figure}
	\centering
	\includegraphics[width=0.98\columnwidth]{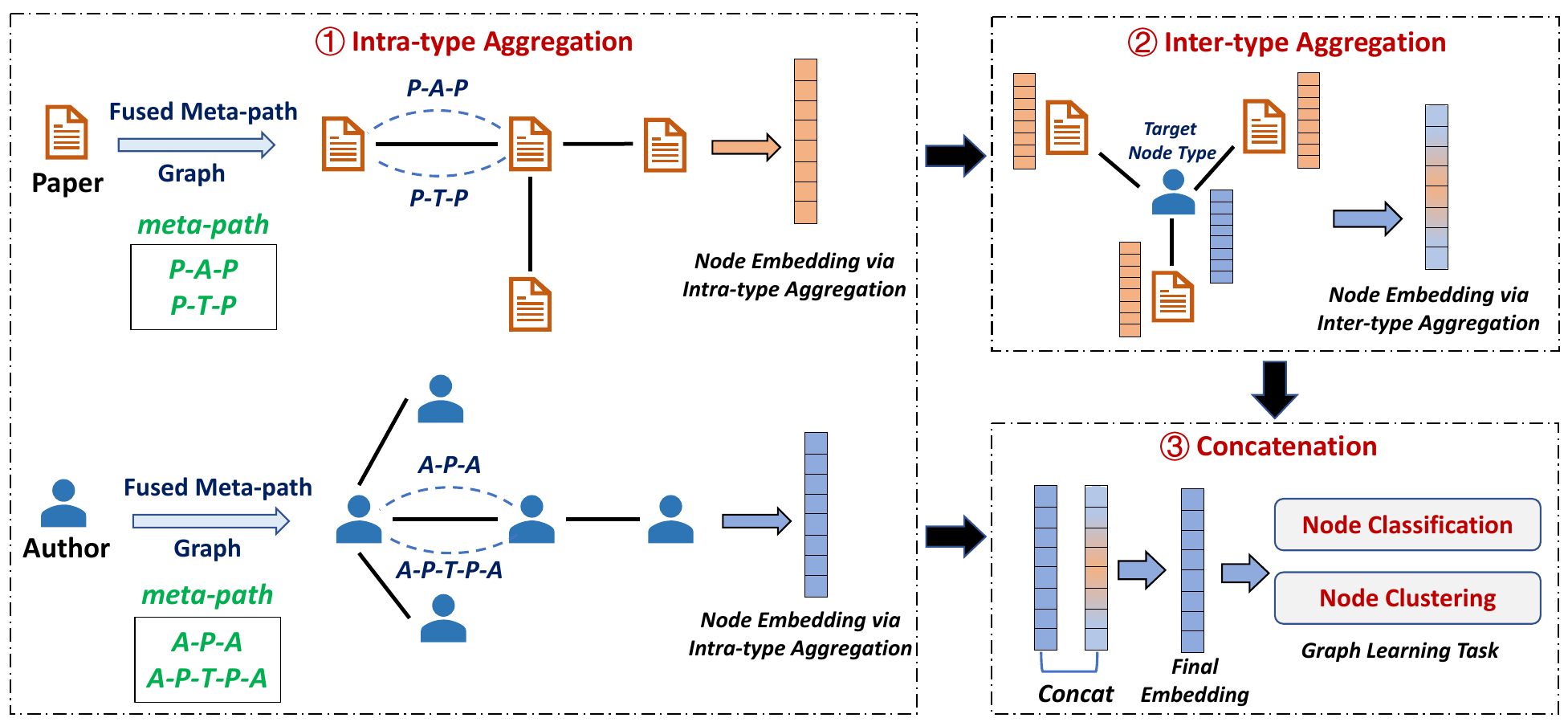}
	\caption
	{
		Visualization of the \mymodel{} process on the DBLP dataset.
	}
	\label{fig:process}
	\vspace{-2ex}
\end{figure}

\subsubsection{Process of \mymodel{}}
The visualization of the \mymodel{} process is demonstrated in Figure~\ref{fig:process}.
We take the DBLP dataset as an example.
For the paper node, the selected meta-paths contain 'P-A-P' and 'P-T-P'.
In the intra-type node aggregation stage, we first construct the fused meta-path graph with the two selected meta-paths.
In the fused meta-path graph of the paper node type, an edge exists between two nodes if these two nodes can be connected by meta-path `P-A-P' or meta-path `P-T-P'.
Then, the intra-type node aggregation is performed in the fused meta-path graph.
Similarly, for the author node, we also perform intra-type aggregation in the fused meta-path graph constructed with `A-P-A' and `A-P-T-P-A'.
Next, we perform the inter-type aggregation in the original graph, where the author node is connected with the paper node.
The initial embeddings of paper and author nodes are obtained through intra-type aggregation. 
Since the target node type is `author', we calculate the embeddings for the author nodes via inter-type aggregation.
Finally, the embeddings obtained through intra-type and inter-type aggregations are concatenated for downstream graph learning tasks.

\subsection{Hyperparameter Sensitivity 
 (RQ5) }

\begin{figure}
  \centering
      \subfigure[DBLP]{
    \includegraphics[width=0.29\columnwidth]{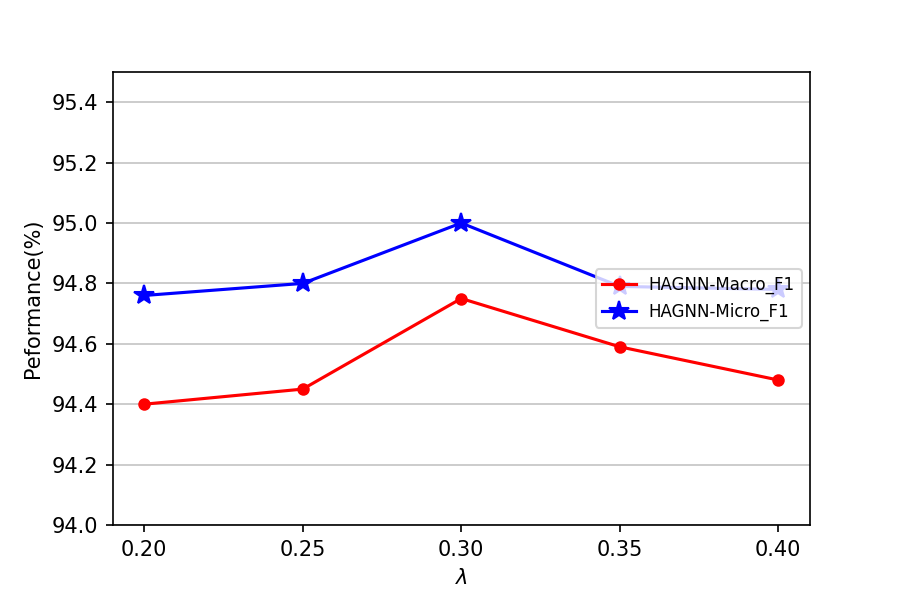}
    }
    \subfigure[IMDB]{
    \includegraphics[width=0.29\columnwidth]{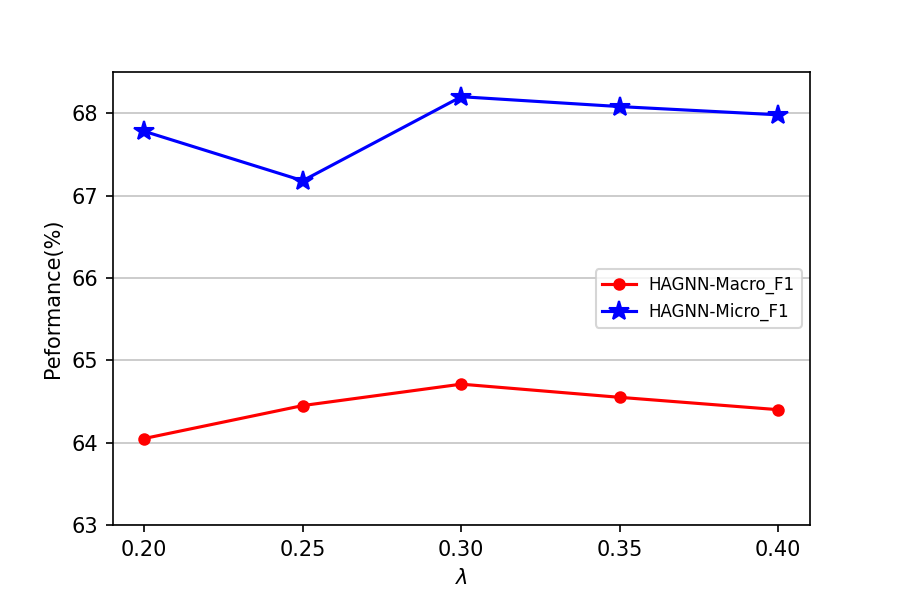}
    }
    \subfigure[PubMed]{
    \includegraphics[width=0.29\columnwidth]{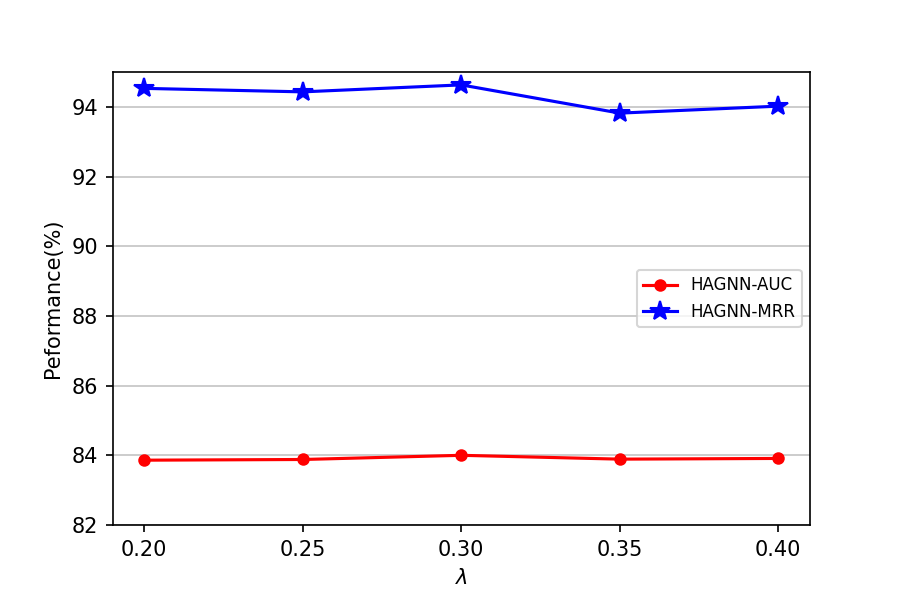}
    }
  \caption{Performance comparison of \mymodel{} under different $\beta$}
  \label{Hyperparameter}
  \vspace{-1ex}
\end{figure}

$\beta$ in Equation~\ref{eq_res} controls how much structural semantic information added into the attention weight.
%
%
We further evaluate the hyperparameter sensitivity on $\beta$.
The value of $\beta$ is selected in $[0.1,0.2,0.3,0.4,0.5]$.
Figure \ref{Hyperparameter} shows the performance comparison of
\mymodel{} with different $\beta$. 
It can be seen that \mymodel{} is very robust to $\beta$, and the performance change is insignificant.
The main reason is that we use the learnable adaptive weight and the structural semantic weight simultaneously, which can increase the robustness of \mymodel{}. 

\subsection{Discussion on Meta-path (RQ6)}
\label{ss:meta-path}

\begin{table}
  \caption{Selected meta-paths in \mymodel{}.}
  \scalebox{0.99}{
  \label{Metapaths selected}
  \centering
  \resizebox{\columnwidth}{!}{
  \begin{tabular}{ccc}
    \toprule
    Datasets     & $\mathcal{T}'$     & Meta-pathes for each type   \\
    \midrule
    \multirow{2}{*}{DBLP} & \multirow{2}{*}{A(Author), P(Paper)} & A:APTPA, APA   \\
    \multirow{2}{*}{} & \multirow{2}{*}{} & P:PAP, PTP \\
     \multirow{2}{*}{IMDB} &\multirow{2}{*}{A(Actor), K(Key)}  & A:AMDMA, AMA   \\
     \multirow{2}{*}{} & \multirow{2}{*}{} & K:KMAMK, KMK \\
    FreeBase &M(Music)  & M:MBOM, MBOBUM   \\
    \multirow{2}{*}{} & \multirow{2}{*}{} & Item:IBI \\
    RCDD &Item ,F & F:FDF   \\
    \midrule
    \multirow{3}{*}{LastFM} &\multirow{3}{*}{U(User), A(Artist), T(Tag)}  &U:UAU   \\
     \multirow{3}{*}{} & \multirow{3}{*}{} & A:AUA \\
     \multirow{3}{*}{} & \multirow{3}{*}{} & T:TAT \\
    \multirow{2}{*}{PubMed} &\multirow{2}{*}{G(Gene), C(Chemical)}  & G:GSDG, GCG   \\
     \multirow{2}{*}{} & \multirow{2}{*}{} & C:CGDC, CSDC \\
      TRD &Poi &Poi:Poi,user,Poi   \\
    \bottomrule
    \end{tabular}}}
\end{table}

\begin{table}
  \caption{Effects of different meta-paths on HAGNN}
  \resizebox{\columnwidth}{!}{
  \label{metapath abli}
  \centering
  \begin{tabular}{cccc}
    \toprule
    \multirow{2}{*}{Dataset}  & \multirow{2}{*}{Meta-pathes}   & \multicolumn{2}{c}{Metrics} \\
     \multirow{2}{*}{}  & \multirow{2}{*}{}   & Macro-F1 & Micro-F1 \\
    \midrule
    \multirow{3}{*}{DBLP} & \textbf{APTPA,APA;PAP,PTP}  &\textbf{95.06$\pm$0.31}&\textbf{95.40$\pm$0.30}  \\
   \multirow{3}{*}{} & APTPA  & 93.53$\pm$0.36  & 94.02$\pm$0.32   \\
   \multirow{3}{*}{} & PAP,PTP,APA  & 94.43$\pm$0.25  & 94.80$\pm$0.24   \\
    \midrule
    \multirow{3}{*}{IMDB} & \textbf{AMDMA,AMA;KMAMK,KMK}  & \textbf{65.57$\pm$0.49} &\textbf{68.62$\pm$0.23}    \\
   \multirow{3}{*}{} & AMDMA,KMAMK  & 64.08$\pm$0.14 &67.21$\pm$0.21   \\
   \multirow{3}{*}{} & AMA,KMK  & 64.59$\pm$0.59 &67.73$\pm$0.63   \\
    \bottomrule
  \end{tabular}}
\end{table}

%
Previous work~\cite{lv2021we} has questioned the effect of meta-paths.
The performance improvement in \mymodel{} verifies that meta-paths are still indispensable for heterogeneous graphs. 
Table~\ref{Metapaths selected} shows the selected meta-paths in \mymodel{}.
To verify the effectiveness of meta-paths, we further conduct experiments with different meta-paths in Table~\ref{metapath abli}.
As shown in Table~\ref{metapath abli}, for the DBLP dataset, meta-paths need to be selected for all types in $\mathcal{T}'$. 
If only one type is selected, the performance will be degraded.
Moreover, the selection of meta-paths should be comprehensive, which can be reflected on the IMDB dataset.

Another question is, how to choose a suitable meta-path.
Meta-paths are important, but arbitrarily choosing meta-paths may cause damage to the model performance. 
%
%
We divide meta-paths into two categories: \textbf{strong-relational} and \textbf{weak-relational meta-paths}. 
Strong-relational meta-path-based graph tend to be sparse, while weak-relational meta-path-based graph are denser.
Take the DBLP dataset as an example. The meta-path `APA' represents the co-author relationship, which is a strong relationship, and the average degree of the `APA'-based-graph is only 3.
%
`APTPA' is a weak relationship because two articles with the same keyword do not indicate that the two authors are closely related. 
The average degree of the corresponding meta-path-based graph is 1232. 

We believe that the sparse strong relationship is difficult to provide enough information in aggregation, while the dense weak relationship provides too much redundant information.
Thus, we can choose the combination of strong relationships and weak relationships. 
Moreover, due to the propose fused meta-path graph and the structural semantic aware aggregation mechanism, \mymodel{} can effectively avoid redundant information and introduce the weight of strong relationships into the learning of the attention weight.
%

    

\subsection{Limitations and Future Works}

Though we provide an analysis about how to select meta-paths, it still requires expert experiences when manually selecting the meta-paths for a given heterogeneous graph, which may be a limitation of \mymodel{}. 
Inspired existing automated graph learning works~\cite{AutoAC, PSP}, we plan to address this issue by designing automated meta-path search methods in future work.
Real-world heterogeneous graphs are often noisy.
Similar to ~\cite{10.1145/3437963.3441734, 10.5555/3524938.3526000, PGA}, we further plan to design robust heterogeneous GNNs for better generalization performance on noisy graph datasets.

\section{CONCLUSION}
In this paper, we proposed a novel hybrid aggregation mechanism for heterogeneous GNNs, which is mainly composed of two stages: the meta-path-based intra-type
aggregation and the meta-path-free inter-type aggregation.
To alleviate the issue of information redundancy in the intra-type aggregation phase, we also designed a novel data structure called fused meta-path graph for meta-path-based aggregation. 
Additionally, we proposed a structural semantic aware aggregation mechanism, which leverages the number of path
instances as the auxiliary aggregation weights.
Extensive experimental results on heterogeneous graph datasets from different real-world domains reveal that \mymodel{} outperforms the existing heterogeneous GNNs in terms of effectiveness and efficiency on node classification and link prediction tasks. 
\section*{Acknowledgment}
This work was supported by the National Natural Science Foundation of China (\#62102177), the Frontier Technology R\&D Program of Jiangsu Province (\#BF2024005), Open Research Projects of Zhejiang Lab (\#2022PG0AB07), and the Collaborative Innovation Center of Novel Software Technology and Industrialization, Jiangsu, China.
%

 \bibliographystyle{IEEEtran}
 \bibliography{reference}

\begin{IEEEbiography}[{\includegraphics[width=1in,height=1.25in,clip,keepaspectratio]{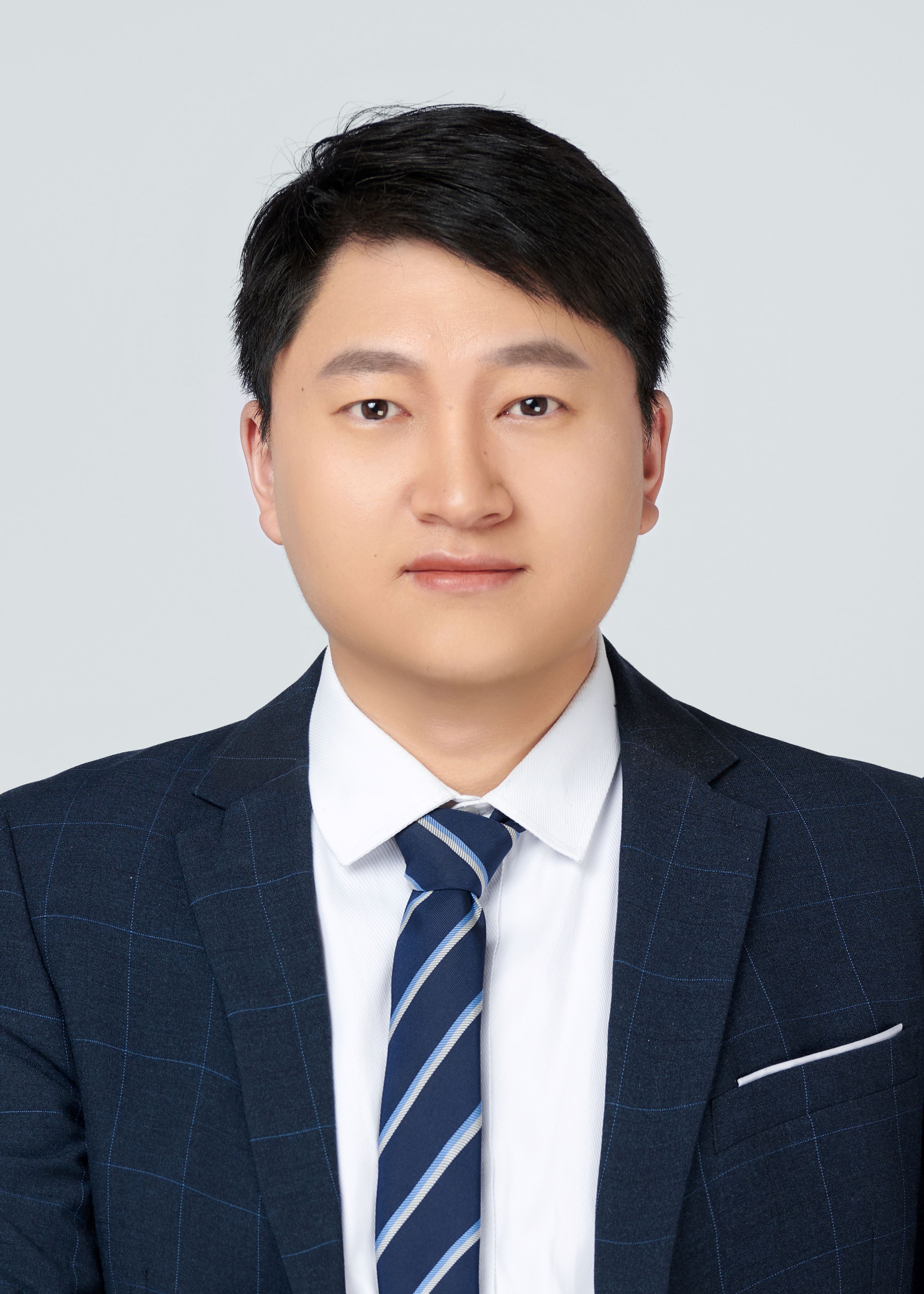}}]{Guanghui Zhu}
is currently an assistant professor in the School of Computer Science, and State Key Laboratory for Novel Software Technology, Nanjing University, China. He received his Ph.D. degree in computer science and technology from Nanjing University. His main research interests include big data intelligent analysis, graph machine learning, and automated machine learning.
\end{IEEEbiography}

\begin{IEEEbiography}[{\includegraphics[width=1in,height=1.25in,clip,keepaspectratio]{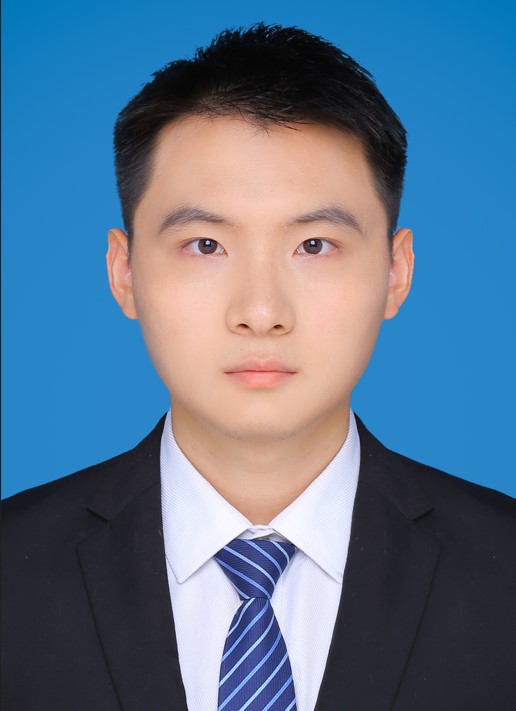}}]{Zhennan Zhu}
is a postgraduate student in the School of Computer Science, Nanjing University, China. He received his BS degree in software engineering from Harbin Institute of Technology, China. His research interests include machine learning, data mining, and graph neutral network.
\end{IEEEbiography}

\begin{IEEEbiography}[{\includegraphics[width=1in,height=1.25in,clip,keepaspectratio]{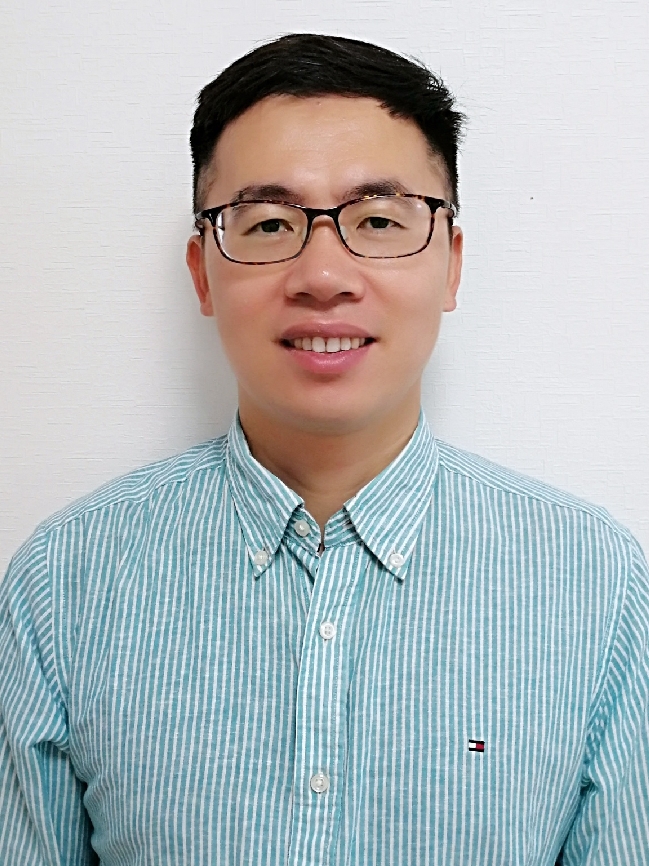}}]{Hongyang Chen}
is a Senior Research Expert with Zhejiang Lab, China. He received the B.S. and M.S. degrees from Southwest Jiaotong University, Chengdu, China, in 2003 and 2006, respectively, and the Ph.D. degree from The University of Tokyo, Tokyo, Japan, in 2011. His research interests include data-driven intelligent systems, graph machine learning, big data mining, and intelligent computing.
\end{IEEEbiography}

\begin{IEEEbiography}[{\includegraphics[width=1in,height=1.25in,clip,keepaspectratio]{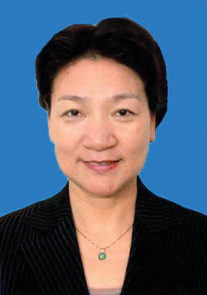}}]{Chunfeng Yuan}
is currently a professor in the School of Computer Science, and State Key Laboratory for Novel Software Technology, Nanjing University, China. She received her bachelor and master degrees in computer science and technology from Nanjing University. Her main research interests include computer architecture, parallel and distributed computing, and information retrieval.
\end{IEEEbiography}

\begin{IEEEbiography}[{\includegraphics[width=1in,height=1.25in,clip,keepaspectratio]{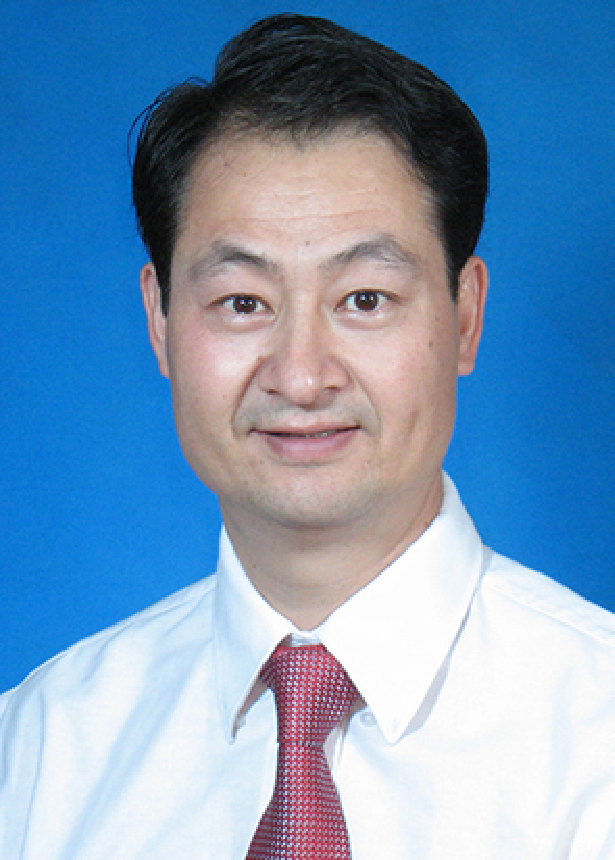}}]{Yihua Huang}
is currently a professor in the School of Computer Science, and State Key Laboratory for Novel Software Technology, Nanjing University, China. He received his bachelor, master and Ph.D. degrees in computer science and technology from Nanjing University. His research interests include parallel and distributed computing, big data parallel processing, big data machine learning algorithm and system.
\end{IEEEbiography}



\vfill

\end{document}